%% file: main.tex
\documentclass[11pt, a4paper, logo, internal]{dm}
\usepackage{graphicx}%
\usepackage{multirow}%
\usepackage{amsmath,amssymb,amsfonts}%
\usepackage{amsthm}%
\usepackage{mathrsfs}%
\usepackage[title]{appendix}%
\usepackage{xcolor}%
\usepackage{textcomp}%
\usepackage{manyfoot}%
\usepackage{booktabs}%
\usepackage{algorithm}%
\usepackage{algorithmicx}%
\usepackage{algpseudocode}%
\usepackage{listings}%
\usepackage{hyperref}
\usepackage{lineno}

\theoremstyle{thmstyleone}%
\theoremstyle{thmstyletwo}%

\theoremstyle{thmstylethree}%
\input{defns}

\def\method{ARIEL}

\title{Advancing AI Research Assistants with Expert-Involved Learning}

\begin{document}

\author[1,8,11]{Tianyu Liu}
\author[1,2,5,11]{Simeng Han}
\author[4,5,11,*]{Hanchen Wang}
\author[3]{Xiao Luo}
\author[5]{Pan Lu}

\author[1,10]{Biqing Zhu}
\author[1,10]{Yuge Wang}
\author[1,10]{Keyi Li}
\author[1,10]{Jiapeng Chen}
\author[1,10]{Rihao Qu}
\author[1,10]{Yufeng Liu}
\author[6,10]{Xinyue Cui}
\author[1,10]{Aviv Yaish}
\author[1,10]{Yuhang Chen}
\author[7,10]{Minsheng Hao}
\author[1,10]{Chuhan Li}
\author[1,10]{Kexing Li}
\author[1,10]{Yinsheng Lu}
\author[1,10]{Xinyu Wei}
\author[1,10]{Qinzhe Xing}
\author[1,10]{Antonia Panescu}
\author[9,10]{Mengbo Wang}
\author[1,10]{Vibha Annaswamy}
\author[1,10]{Alicia Sanchez}
\author[1,10]{Jack Cloherty}

\author[1]{Arman Cohan}

\author[1]{Hua Xu}

\author[1]{Mark Gerstein}

\author[5]{James Zou}

\author[1,*]{Hongyu Zhao}

\affil[1]{Yale University}

\affil[2]{Google DeepMind}

\affil[3]{University of Wisconsin–Madison}

\affil[4]{Genentech}

\affil[5]{Stanford University}

\affil[6]{Cornell University}

\affil[7]{Harvard University}

\affil[8]{Broad Institute of MIT and Harvard}

\affil[9]{Purdue University}

\affil[10]{These authors contribute equally to this project as human experts}

\affil[11]{These authors contributed equally to this work. Contact email: tianyu.liu@yale.edu}
\affil[*]{Corresponding Authors.}




\begin{abstract}
Large language models (LLMs) and large multimodal models (LMMs) promise to accelerate biomedical discovery, yet their reliability remains unclear. We introduce ARIEL (AI Research Assistant for Expert-in-the-Loop Learning), an open-source evaluation and optimization framework that pairs a curated multimodal biomedical corpus with expert-vetted tasks to probe two capabilities: full-length article summarization and fine-grained figure interpretation. Using uniform protocols and blinded PhD-level evaluation, we find that state-of-the-art models generate fluent but incomplete summaries, whereas LMMs struggle with detailed visual reasoning. We later observe that prompt engineering and lightweight fine-tuning substantially improve textual coverage, and a compute-scaled inference strategy enhances visual question answering. We build an ARIEL agent that integrates textual and visual cues, and we show it can propose testable mechanistic hypotheses validated by the feedback from human experts. ARIEL can also work with human experts to build ``expert-in-the-loop" framework and correct issues in experts' answers. Overall, ARIEL delineates current strengths and limitations of foundation models, and provides a reproducible platform for advancing trustworthy AI in biomedicine.
\end{abstract}




\maketitle
\section{Introduction}
\input{section_folder/introduction}

\section{Results}
\input{section_folder/Results}

\section{Discussion}
\input{section_folder/Discussion}

\section{Methods}
\input{section_folder/Methods}

\bibliographystyle{unsrt}
\bibliography{sn-bibliography}

\appendix
\input{section_folder/Appendix}

\end{document}

%% file: defns.tex
\usepackage{mathtools}
\usepackage[dvipsnames]{xcolor}
\usepackage[numbers, sort&compress, round]{natbib}
\usepackage{booktabs}
\usepackage{graphicx}
\usepackage{xfrac}
\usepackage{bbm}
\usepackage{changes}
\usepackage{rotating}
\usepackage{wrapfig}

\usepackage[most]{tcolorbox}
\usepackage{xparse}
\usepackage{adjustbox}
\usepackage{xspace}
\usepackage{changepage}
\usepackage{enumitem}
\usepackage{pifont}
\usepackage{ulem}
\usepackage{tocloft}
\usepackage[toc]{multitoc}
\usepackage{etoc}
\usepackage{dsfont}
\usepackage{dm-colors}
\usepackage{multirow}
\usepackage{minted}
\usepackage{caption}
\usepackage{soul}
\usepackage{float}
\usepackage{svg}
\usepackage{adjustbox}
\usepackage{setspace}
\svgsetup{inkscapelatex=false}
\usepackage{silence}  
\newcommand{\hide}[1]{}

\newcommand{\method}{SpatialAgent\,}

\pdftrailerid{redacted}

%% file: section_folder/introduction.tex
The success of Foundation Models (FMs) \cite{bommasani2021opportunities}, including Large Language Models (LLMs) and Large Multi-Modal Models (LMMs), has been demonstrated across various applications, garnering widespread public interest in research focused on these models \cite{singhal2023large, acosta2022multimodal, tu2024towards, han2025ateb}. A very important target of an FM is to understand various modalities and model them to enable multimodal reasoning \cite{wei2022chain,zhang2023multimodal} in a human-like manner. These human skills give us the ability to understand what is being conveyed in a multimodal context, while at the same time being able to externalize it with our linguistic system for processing and refining the content of the text \cite{johnson2015logic}. In this work, we investigate how we can utilize FMs to help humans summarize long research papers as well as understand multimodal scientific context. 

There are various long documents in the biomedical domain, including examination reports, patient case data, research manuscripts, and others \cite{chatterjee2021informed}. These documents may also include figures as extra information to help us understand the corresponding content. However, due to the explosion of biomedical data nowadays, physicians and researchers are struggling more and more to understand and process such documents in biomedical areas \cite{wang2019big}. Also, the cost of hiring additional human resources to help process the data is high and therefore not optimal for the biomedical field \cite{wijeysundera2012techniques, van2024adapted}. At the same time, there has been evidence \cite{stacey2003complex, bystrom1995task, wilfredo2013acquiring} to show that humans tend to engage in complicated reasoning processes and expand their knowledge when answering realistic questions. For example, it is common to use multiple examination results to determine the condition \cite{sanders2010every} during the diagnosis from a physician. Therefore, how to measure and optimize the reasoning capacity of FMs for complex multimodal data is an important and practical research problem \cite{han2024hybridmind}. If an FM shows an excellent ability to summarize long texts as well as complicated figures with different panels, it will become a powerful research assistant in this field, which can shorten the time for physicians to make correct diagnoses and for researchers to understand the work of their peers.
Previous research has investigated the ability of LLMs to summarize non-medical documents including news \cite{zhang2024benchmarking}, codes \cite{haldar2024analyzing}, and reviewer comments \cite{li2024chatcite, jin2024comprehensive}. Research has also been conducted to evaluate LLMs for summarizing short medical documents including clinical texts \cite{van2024adapted} and medical evidence \cite{tang2023evaluating}. Meanwhile, the ability of LMMs to summarize scientific figures has also been explored \cite{wang2024scientific, roberts2024scifibench}. These papers argue that some advanced models such as GPT-4 can outperform human experts in highlighting the abstract of certain documents or figures, while all models have their limitations due to either hallucinations or out-of-domain queries. Furthermore, most of the recent LLM or LMM evaluations have focused on comparisons among models but ignored the role of humans in evaluations. However, humans are very important references in evaluating the utilities of FMs for their purpose, and understanding the differences between FMs and humans can improve the reliability of these FMs for scientific research.

As far as we know, comprehensive human-aware evaluations of LMMs for understanding long biomedical documents and biomedical figures remain unexplored. Although there are several LMM benchmarks on low-level knowledge quizzes, they cannot be applied to our problem due to the complexity of biomedical resources~\cite{xiao2024logicvista, yue2024mmmu,sun2024pathmmu}. In addition, due to the rapid development of LMMs, we expect more up-to-date evaluations and benchmarks in this time-sensitive domain~\cite{white2024livebench}. Most of the existing baselines also do not include human participators as an important metric. Researchers need to get fair and unbiased results carefully to avoid leakage of the test set. Furthermore, inspired by previous work focusing on generating scientific hypotheses from text-level information \cite{qi2024large, gottweis2025towards}, we are interested in understanding how multimodal biomedical data can help generate new research questions to accelerate scientific progress. Therefore, our LMM evaluations and improvement are challenging and meaningful, which can fill a gap in this advanced area.

Here we present a novel dataset accompanied by an analytical framework, named \method{}, for evaluating and improving the understanding ability of LLMs and LMMs in the biomedical domain. Here, we focus on two basic tasks, i.e., long manuscript summarization (mainly related to COVID-19 and healthcare) \textcolor{red}{and complicated figure comprehension (mainly related to multiomics, neuroscience, and pathology) and provide several strategies for performance enhancement.} To make a fair and comprehensive comparison, we manually constructed an open-source biomedical article summarization dataset as well as a biomedical figure understanding dataset. Our benchmark, based on \method{}, not only systematically evaluates the performance of representative open- and closed-source LLMs and LMMs on both tasks, but also offers human-aware evaluations from experts in the biomedical field. \textcolor{red}{By analyzing the outputs and experts' answers, we summarize the strengths and weaknesses of each model type, which enables us to explore how to leverage FMs to facilitate scientific inquiries. We also test the performance of \method{} in generating biomedical hypotheses with different frameworks. The evaluations of these tasks cover both computational metrics and human evaluation results. Overall, we believe that our multi-functional framework can advance the applications of LLMs and LMMs in the domain of biomedicine and healthcare.}

%% file: section_folder/Results.tex
\begin{figure}
    \centering
    \includegraphics[width=\linewidth]{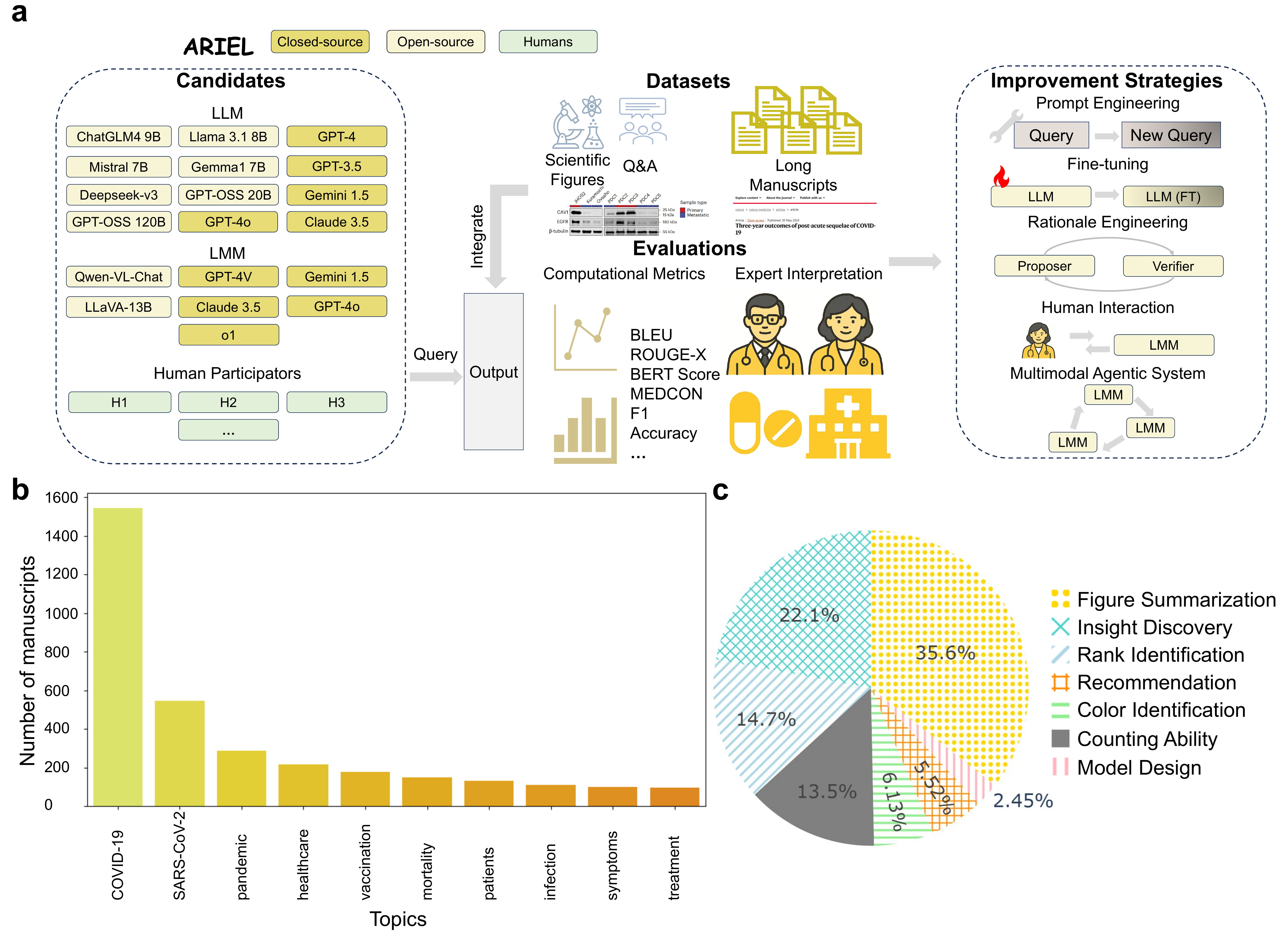}
    \caption{Landscape of \method{}. (a) We designed a framework supporting 1. the evaluation of both open-source and closed-source LLMs and LMMs; and 2. the utilization of LLMs and LMMs for helping human researchers resolve both the text summarization task and the figure understanding task with new datasets. We also introduced numerical evaluation metrics and human-aware evaluation metrics for both tasks. Our framework also contains solutions for improving model performances, which are listed in the panel. (b) Number of papers ($n=2571$) with top 10 topics in datasets used for biomedical document summarization based on number ranking. (c) Proportion of questions ($n=163$) in different categories for scientific figure understanding.}
    \label{fig: method overview}
\end{figure}

\textbf{Overview of \method{}} To systemically assess the abilities of FMs in understanding research papers, we constructed two datasets: one from manuscripts and another from figures. Data were extracted from PubMed \cite{white2020pubmed} and recently published top-tier journals, with no overlap with the cut-off dates of model training sets. We then developed a general pipeline for generating and automatically evaluating outputs from both closed- and open-source FMs on these datasets (\textbf{Methods}).
For text summarization, a language model receives a manuscript and a prompt as input; for figure understanding, a multimodal model takes the figure, its description, and a prompt. In addition to automated metrics, we conducted a human evaluation (Human-Eval) with 20 doctoral-level biomedical researchers who judged model outputs and their sources. We then analyzed all experimental results to identify the gaps between current FM capabilities and the requirements of an effective AI research assistant.
Finally, we propose initial strategies to leverage FMs for summarizing documents, interpreting figures, and generating hypotheses. An overview of the \method{} framework is shown in Figure~\ref{fig: method overview}.

\textbf{Datasets.} 
All questions in our datasets are open-ended. For document summarization, we collected 2571 manuscripts via the PubMed API. Compared with existing benchmarks \cite{chen2025benchmarking, wang2022squality, bai2023longbench}, \method{} is both domain-specific and more rigorous, reflecting the complexity of scientific language. We posed open-ended prompts rather than standard QA to increase difficulty. Manuscripts were split into training and test sets, with validation sets derived automatically \cite{zheng2024llamafactory}. Topic analysis (Figure~\ref{fig: method overview} (b)) shows dominance of COVID-19 and SARS-CoV-2, followed by healthcare and treatment, reflecting a focus on respiratory epidemics.


For figure understanding, we curated \textcolor{red}{20 high-quality figures from peer-reviewed publications \cite{roberts2024scifibench, xiao2024logicvista}}, paired with 163 open-ended questions spanning summarization, color identification, model design, rank identification, recommendation, and insight discovery (Figure~\ref{fig: method overview} (c)). Compared with prior benchmarks \cite{roberts2024scifibench, xiao2024logicvista}, our dataset prioritizes low noise and questions requiring reasoning. Human evaluation (Human-Eval) included all figures and a random sample of 10 manuscripts.


\begin{figure}
    \centering
    \includegraphics[width=1\linewidth]{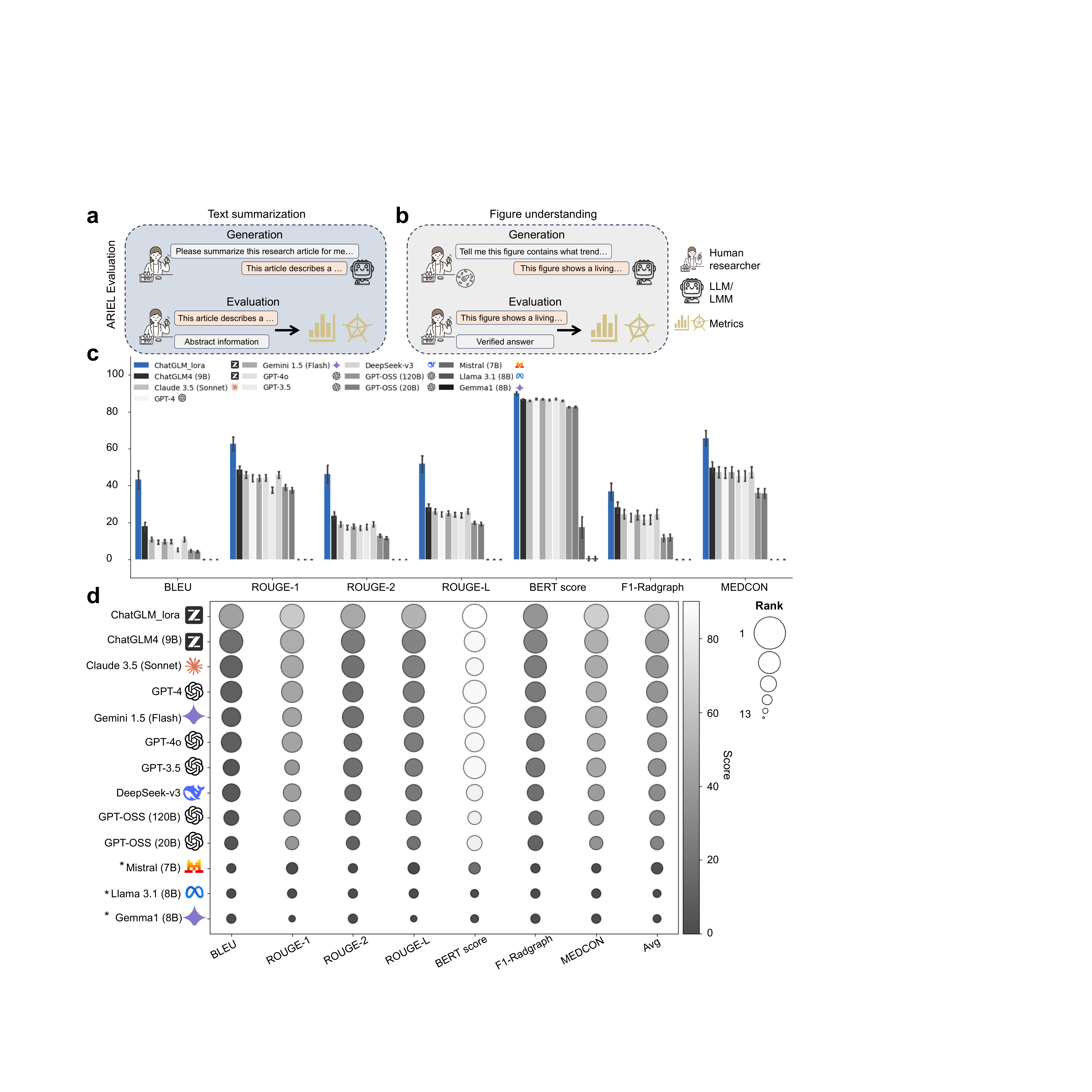}
    \caption{Evaluation pipelines and results of text summarization task. We report the average and standard error in all bar plots. (a) Pipeline of evaluation framework for text summarization. (b) Pipeline of evaluation framework for figure understanding (c) Bar plots with standard error of model performances across different evaluation metrics. (d) Bubble plot of averaged model performances (bubble color) and ranks (bubble size) by different evaluation metrics. Models with averaged scores lower than 1.0 are annotated with *.}
    \label{fig: llm evaluation metric}
\end{figure}

\textbf{Large-scale evaluations of long document summarization identify fine-tuned models as better performers.} 
We benchmarked both open-source LLMs (ChatGLM4 (9B) \cite{glm2024chatglm}, Gemma1 (8B) \cite{team2024gemma}, Llama 3.1 (8B) \cite{touvron2023llama}, Mistral (7B) \cite{jiang2023mistral}, Deepseek-v3 \cite{liu2024deepseek}, GPT-OSS (20B) \cite{gptoss2025}, GPT-OSS (120B) \cite{gptoss2025}) and closed-source LLMs (GPT-3.5 \cite{papergpt3}, GPT-4 \cite{achiam2023gpt}, GPT-4o \cite{hurst2024gpt}, Gemini-Flash-1.5 \cite{team2023gemini}, and Claude-Sonnet-3.5 \cite{antrop2024}). Performance was assessed with seven established metrics (BLEU, ROUGE-1/2/L, BERTScore, F1-Radgraph, MEDCON)\cite{van2024adapted, yim2023aci,papineni2002bleu, lin2004rouge, zhangbertscore, jain1radgraph, yim2023aci}. All of the metrics range from 0 to 100, and a higher value means a better output (Figure \ref{fig: llm evaluation metric} (a), \textbf{Methods}). Figure \ref{fig: llm evaluation metric} (b) shows the pipeline for LMM evaluation, which will be discussed later. To test whether task-specific adaptation improves performance, we fine-tuned the best-performing open-source model, ChatGLM4, using Low-Rank Adaptation (LoRA \cite{hulora}, as a well-known parameter-efficient fine-tuning method).

Based on Figure \ref{fig: llm evaluation metric} (c), we analyzed the distribution and variance of scores across all evaluation metrics. Overall, open-source models operating in zero-shot inference mode exhibited lower performance \textcolor{red}{by mean scores across the seven automated metrics in Figure \ref{fig: llm evaluation metric} (c)–(d).}. Even advanced open-source systems, such as DeepSeek-V3 and the GPT-OSS series, achieved comparatively \textcolor{red}{modest average normalized scores across the seven metrics}, while models like Gemma1, Llama 3.1, and Mistral often failed to generate meaningful outputs \textcolor{red}{as reflected by near-zero scores on ROUGE/BERTScore}. These shortcomings are likely attributable to data constraints and limited context length. By contrast, ChatGLM4, Claude-Sonnet-3.5, and GPT-4 consistently ranked among the top three models for long-document summarization, aligning with prior studies on the use of LLMs as reviewers \cite{zhou-etal-2024-llm}. Notably, ChatGLM4 surpassed other open-source models across \textcolor{red}{metrics including BLEU, ROUGE-1/2/L, BERTScore, F1-Radgraph, and MEDCON ( Figure \ref{fig: llm evaluation metric} (c)),} and demonstrated competitive performance relative to closed-source systems. To further explore its capacity, we applied different prompting strategies and fine-tuned the model (ChatGLM\_lora). As shown in Extended Data Figure \ref{supfig: prompt eng text task}, prompt engineering methods including meta prompting \cite{suzgun2024meta} and chain-of-thought reasoning (CoT) \cite{wei2022chain} had minimal impact on performance. However, fine-tuning substantially improved outcomes \textcolor{red}{by increasing the mean score across the seven metrics (Figure \ref{fig: llm evaluation metric} (c)) and improving average rank (Figure \ref{fig: llm evaluation metric} (d)).}, with the adapted model outperforming all closed-source LLMs. This result suggests that domain-specific training can significantly enhance summarization quality and that fine-tuned, smaller-scale models can achieve strong task-specific performance. In Figure \ref{fig: llm evaluation metric} (d), we display the average score of each metric for all the models, and the results of both performances and ranks are consistent with conclusions from Figure \ref{fig: llm evaluation metric} (c).

\begin{figure}
    \centering
    \includegraphics[width=1\linewidth]{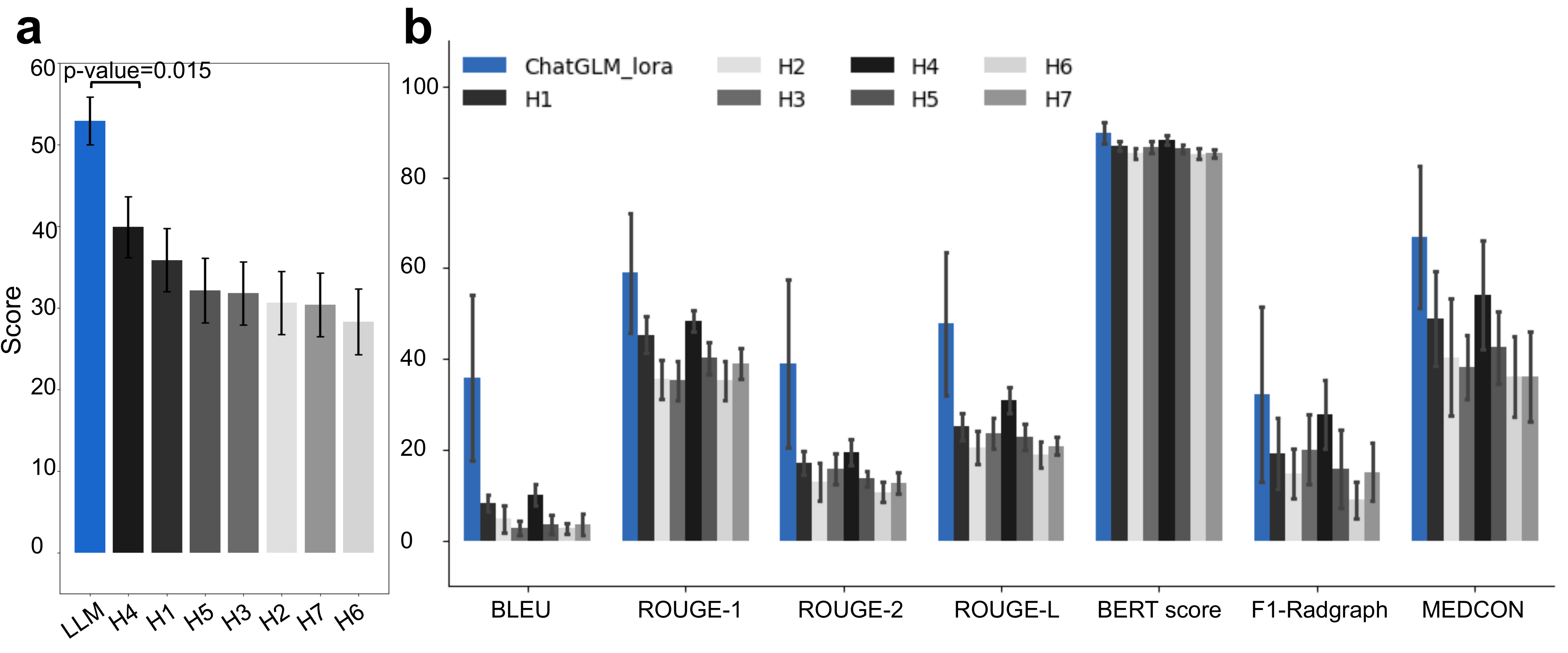}
    \caption{Comparisons between human participators and LLMs for the text summarization task. We report the average and standard error in all bar plots. (a) Bar plots of averaged scores and standard errors across all metrics by different participators and ChatGLM\_lora. (b) Bar plots of participators' performances by different evaluation metrics.}
    \label{fig: llm human compare text summ}
\end{figure}

\textbf{Interpretation of the human-generated results of long document summarization.} 
We collected expert feedback on abstract generation for selected scientific papers, denoted as human-generated results. The participants were highly trained specialists in biology, and their evaluations provide critical insights into the divergence between human and LLM preferences when summarizing long documents. \textcolor{red}{As described in the Dataset section, we randomly sampled ten papers from the test set (the sample size was chosen to balance expert burden with an unbiased assessment), and seven participants (H1–H7) provided feedback.} As shown in Figure \ref{fig: llm human compare text summ} (a), the fine-tuned LLM, ChatGLM\_lora, can still outperform the best human participator H4 significantly (52.95 v.s. 39.93), supported by the p-value computed based on the two-sided Wilcoxon rank-sum test (p-value=0.015) between the computational metrics. Moreover, we demonstrated the performance comparison stratified by different metrics, shown in Figure \ref{fig: llm human compare text summ} (b). According to this figure, ChatGLM\_lora outperformed other human participators in all of the metrics, which demonstrates that the improvement of fine-tuned LLM is comprehensive. Moreover, we compared the time cost and monetary cost brought by humans and models. Here the cost of machines is computed based on running inference with the fine-tuned ChatGLM model with a standard NVIDIA H100 GPU (per hour price: 0.37\$) \cite{lambdagpucost}. The cost of human participators is computed based on averaging time across different participators based on their reports, and the per-hour price is estimated based on 100,000\$ as annual salary (per hour price: 48.0\$) \cite{incomecost}. According to Extended Data Figure \ref{supfig: time cost compare} (a), using LLM can reduce both time cost and monetary cost. Overall, these results indicate that fine-tuned LLMs can generate abstracts more closely aligned with original texts than human experts from the perspective of context similarity, but fine-tuned LLMs also show an increase in output variability.

Furthermore, to compare the differences between abstracts from human participators and LLMs in detail, we show an example of one abstract in Extended Data Figure \ref{supfig: text sup info}. By comparing the original abstract and the human-generated abstract, we found that human-generated outputs have more unique information, highlighted by the blue fonts. The original abstract provides detailed epidemiological data, exact hazard ratios, and a more formal structure. The human-generated report offers a concise overview, underscores the need for broader research, and specifically mentions the lack of a significant finding regarding certain ART regimens. Meanwhile, they also have similarities, as both paragraphs describe the same nationwide South Korean study that investigated outcomes of COVID-19 in People Living with HIV (PLWH) compared to People Living without HIV (PLWoH). They agree that although the crude mortality rate was higher for PLWH, the adjusted difference in mortality was not statistically significant, while the risk of severe or critical illness was clearly elevated. By comparing the original abstract and the LLM-generated abstract, we found that there are substantial numerical inconsistencies between them regarding total cases, proportions of PLWH, comorbidity rates, and mortality. We suspect that these inconsistencies may be due to the wrong numerical understanding made by LLMs when generating the outputs, which might preclude the readers from reaching the correct conclusion. Nevertheless, both paragraphs can describe a nationwide epidemiological study in South Korea over the same time frame, concluding that PLWH is at a higher risk of poor COVID-19 outcomes (either higher incidence or greater severity) and requires dedicated strategies to reduce morbidity and mortality. Therefore, LLMs have advantages in generating an abstract with a more well-structured format and its conclusion is closer to the original description, and human participators have advantages in summarizing the high-level concepts and providing new ideas for including other aspects in the abstract. We also note that the LLMs have some problems with generating data support, so the user needs to analyze the statistics in the output in more detail when using LLMs and make necessary corrections. In Extended Data Figure \ref{supfig: text enrich example}, we show an example of a generated abstract with more textual information, which contains less inconsistent information and is more reliable.

\textcolor{red}{Finally, we examined the divergence between fine-tuned LLMs and human outputs by projecting both into an embedding space using the OpenAI embedding model (text-embedding-large). Cosine similarity scores across all selected articles are presented in Extended Data Figure \ref{supfig:sim info} (a). The results indicate a high degree of alignment, with nearly all similarity values exceeding 0.8, suggesting that LLM-generated content captures contextual information comparable to that produced by humans.}

\textcolor{red}{Meanwhile, we also invited 10 human experts to evaluate the quality of abstracts from three different approaches (original paper (Abstract), LLM-generated (LLM), and human-reported (Human)) from five different dimensions (Completeness, Relevance, Conciseness, Coherence, and Clarity. Details can be found in the Methods section). The study is conducted in a blinded format, allowing it to assess the quality of different types of abstracts from the perspective of human researchers, rather than focusing on numerical similarity. According to Extended Data Figures \ref{supfig: llm generated score} (a) and (b), we visualize the scores by metrics and papers, and find that human experts believe that the abstracts from original papers have the highest average quality, followed by the abstracts from human experts and LLMs. The difference is also significantly validated by the p-value computed with the two-sided Wilcoxon Rank-sum test. Therefore, human researchers still prefer reading the abstract extracted from the original manuscript, which might contain more informative messages. Moreover, abstracts from LLMs have higher completeness scores compared with the results from humans, which aligns with our previous study based on computational metrics and a similarity study. However, abstracts from humans have higher scores in Relevance, Conciseness, Coherence, and Clarify. Therefore, by combining the comparison of computational metrics with evaluations conducted by human experts, we have concluded that while LLM-generated abstracts contain more information from the original articles, human-written abstracts are more readily accepted by human researchers. Therefore, future research on abstract generation should focus on producing content that better aligns with human preferences. There is also potential to develop better datasets in summarization model training.}

\begin{figure}
    \centering
    \includegraphics[width=1\linewidth]{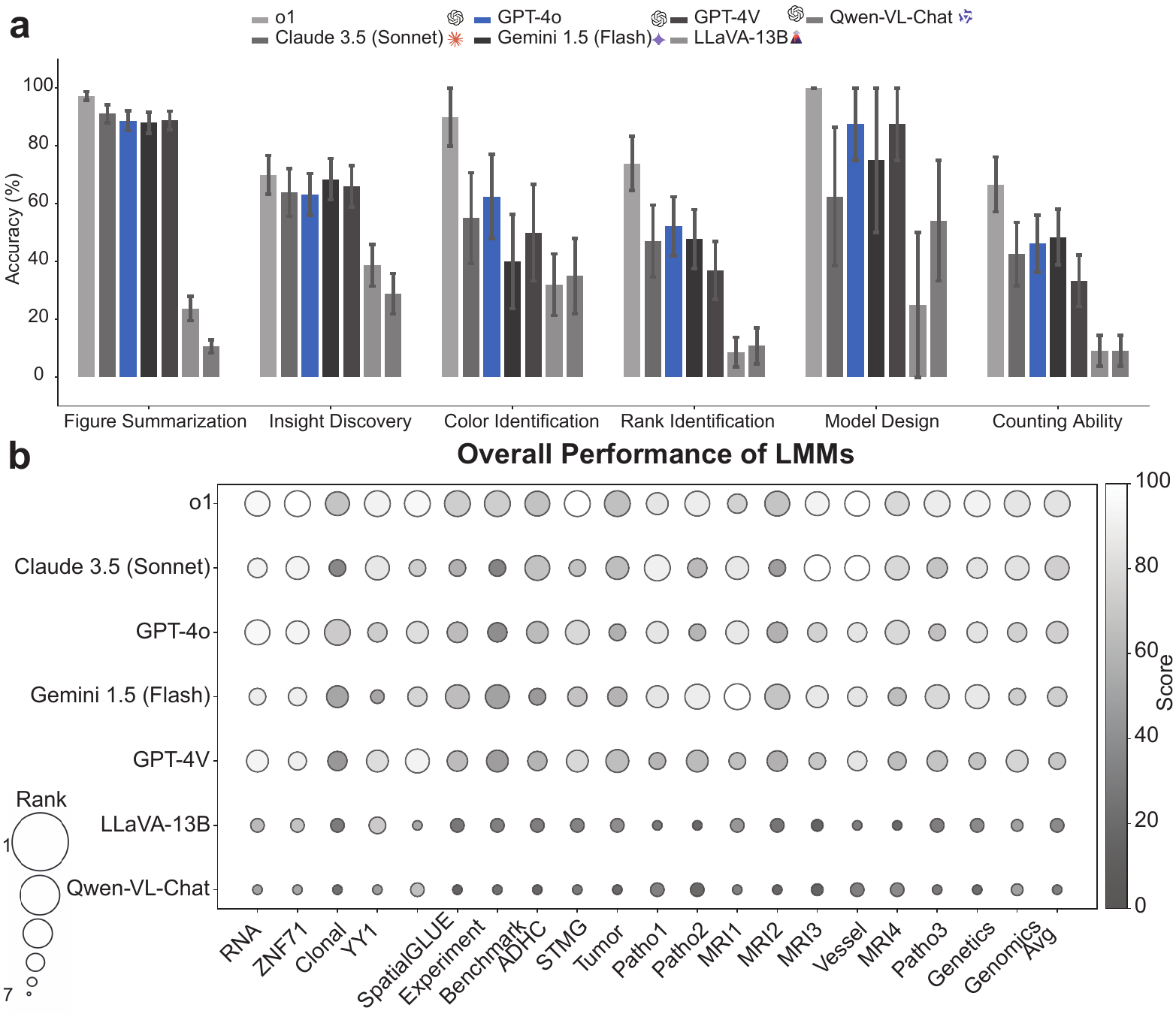}
    \caption{Evaluations of LMM's ability to understand scientific figures. We report the average and standard error in all bar plots. (a) Bar plot of accuracy and standard error from different LMMs' performances by the categories of questions. (c) Bubble plot of average performances (bubble color) and ranks (bubble size) of different questions by datasets for all selected models.}
    \label{fig: lmm evaluation metric}
\end{figure}

\textbf{Without test-time computation scaling, current LMMs fail to fully understand scientific figures in research papers.} In this section, we assess the capacity of LMMs to interpret scientific figures. Human interpretation of such figures typically involves understanding both the overall context and the detailed content of individual panels, as well as deriving potential insights or biological discoveries. To evaluate these dimensions, we designed a set of questions targeting different levels of comprehension and included both open-source models (Qwen-VL-Chat \cite{bai2023qwen}, LLaVA-13B \cite{liu2023visual}) and closed-source models (Gemini-Flash-1.5 \cite{team2023gemini}, GPT-4V, GPT-4o, Claude-Sonnet-3.5 \cite{antrop2024}, and o1 \cite{jaech2024openai}). The o1 model, in particular, leverages test-time computation \cite{snell2024scaling} to improve reasoning by employing search strategies to identify the most accurate reasoning path when generating responses.

According to Figure \ref{fig: lmm evaluation metric} (a), the outputs of LMMs exhibited substantial variability across different question types, as illustrated by the plots in each panel. Among the evaluated models, o1 consistently achieved the highest \textcolor{red}{mean accuracy} across all question categories, whereas open-source models performed markedly worse \textcolor{red}{in mean accuracy} than closed-source counterparts. Notably, all models showed considerable variance on tasks involving insight discovery and color identification, indicating persistent limitations in generating novel biological inferences and distinguishing color differences. Figure \ref{fig: lmm evaluation metric} (b) presents the average scores across datasets for each method. Here, o1 demonstrated robust \textcolor{red}{mean accuracy} across categories, underscoring the utility of test-time computation in enhancing visual reasoning. Nonetheless, o1 also displayed high variance in tasks including Insight Discovery, Rank Identification, and Counting. \textcolor{red}{GPT-4o performed strongly on datasets involving RNA-seq and vessel structure analyses (RNA (95.83), ZNF71 (100), and Vessel (100)) but struggled with figures focused on benchmarking analyses (Benchmark (68.57)).} By contrast, GPT-4V exhibited notable strength in interpreting model design principles, particularly in the SpatialGLUE dataset. Overall, these findings highlight the need for more advanced models with improved capacity to interpret scientific figures across diverse domains and task types.

Finally, inspired by \cite{kadavath2022language}, we investigated whether LMMs are capable of producing reliable self-assessments of confidence. This analysis is critical for understanding the internal logic guiding model-generated content. Specifically, we asked o1, GPT-4V, and Claude-Sonnet-3.5 to assign confidence scores (ranging from 0 to 1, with higher values indicating greater confidence) to their answers across all questions. As shown in Extended Data Figure \ref{supfig: confidence plot}, neither GPT-4V nor Claude-Sonnet-3.5 produced meaningful confidence estimates, as their assigned scores were not significantly correlated with \textcolor{red}{accuracy} (Spearman correlation, p = 0.11 and p = 0.14, respectively). In contrast, o1 exhibited a significant positive correlation between confidence and accuracy (p = 0.006), although inconsistencies remained, with the model assigning divergent confidence levels to responses with identical scores. Overall, these findings raise concerns about the reliability of confidence estimation in current LMMs and underscore the need for careful human oversight when employing them as assistants for scientific figure interpretation.

\begin{figure}
    \centering
    \includegraphics[width=1\linewidth]{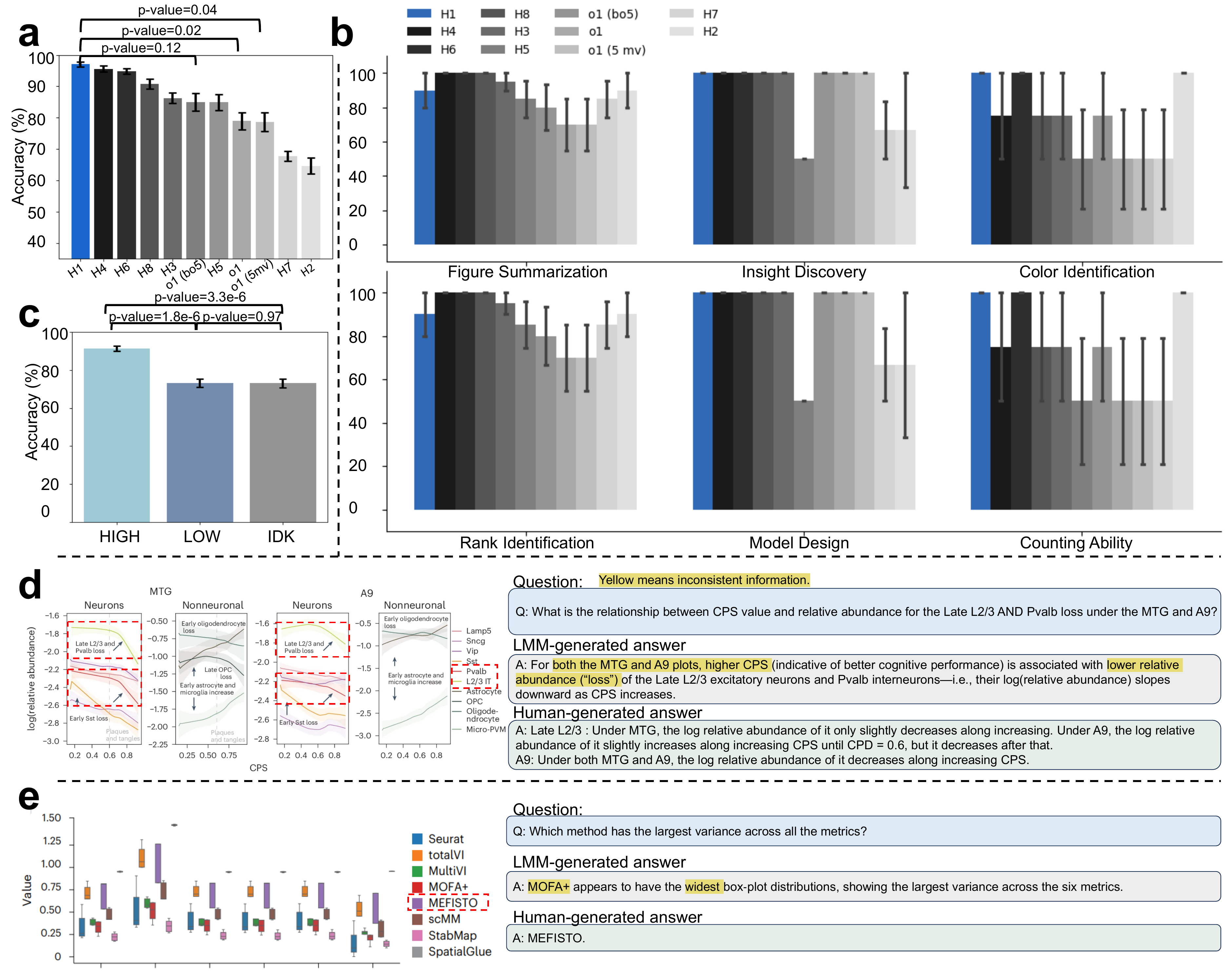}
    \caption{Comparisons between human participators and LMMs for the scientific figure understanding task. We report average and standard error in all bar plots. (a) Bar plot of average scores across all datasets by different participators. (b) Bar plots of accuracy and standard errors from participators' performances by different problem categories. (c) Bar plot of accuracy by confidence levels provided by different human participators. We have three options for humans to choose. (d) The first case study is based on the original question, LLM-generated answer, and Human-generated answer. (e) The second case study is based on the original question, LLM-generated answer, and Human-generated answer.}
    \label{fig: lmm human evaluation metric}
\end{figure}

\textbf{Interpretation of the Human-Eval results of figure understanding.} In this section, we compare the performance of human participants and LMMs on the figure understanding task, highlighting both similarities and differences in their outputs. To approximate the upper bound of LMM performance, we introduced two additional baselines for o1: the five-majority-vote variant (\textit{o1 (5 mv)}) and the best-of-five variant (\textit{o1 (bo5)}), the latter option serving as an oracle. We also had eight human experts (H1–H8) in the evaluation. \textcolor{red}{According to Figure \ref{fig: lmm human evaluation metric} (a), the performer with the best average accuracy in answering figure-related questions is still human, while LMMs with test-time computation with three different variations can only surpass two humans \textcolor{red}{in mean accuracy}. This observation still holds for the 10 additional images introduced in the revision, according to Extended Data Figure \ref{supfig: additional image} (human expert's accuracy is 88\%, while o1's accuracy is 83\%).} Moreover, major voting improves the accuracy of the LMM and reduces the difference in performance compared with the best human participators. This conclusion is supported by the p-value difference of the two-sided Wilcoxon rank-sum test. However, major voting consumes more tokens and therefore leads to higher costs. Therefore, we also explored how to utilize rationale engineering to improve the answer accuracy of LMM in the next section. In Figure \ref{fig: lmm human evaluation metric} (b), we visualize the differences between the performances of LMMs and humans by categories. We found that \textit{o1 (5 mv)} ranks higher than o1 in figure summarization and counting ability apparently, the LLMs might have more randomness in generating answers related to these tasks. Moreover, in all of the categories, the best performer is human, but the human participators also have in-group variation across different categories. However, according to Extended Data Figure \ref{supfig: time cost compare} (b), LMMs still have advantages in reducing both the time cost and monetary cost for this task. Here the monetary cost of LMMs is computed based on the token price of API calling \cite{openaiapitoken}. These findings indicate that integrating human feedback remains valuable for improving LMM performance. However, to maximize benefits, careful selection of reference participants is necessary, prioritizing individuals with higher accuracy to ensure the reliability of training or evaluation data.

When designing the questionnaire, we also asked human participators to provide confidence evaluation for their outputs, stratified by I don't know (in short, idk, which can be interpreted as the lowest certainty or random guess), low certainty (low), and high certainty (high). According to Figure \ref{fig: lmm human evaluation metric} (d), responses in the high-confidence group achieved significantly higher accuracy than those in the other groups, with statistical significance confirmed by the Mann–Whitney U test. These results suggest that human self-assessed confidence is more reliable than LLM-generated confidence and may serve as a useful signal for calibrating LLM outputs.

We further conducted error analysis of generated answers, with error types defined following prior studies \cite{hua2025inductionbench,miner2024scheherazade}. Figure \ref{fig: lmm human evaluation metric} (d) illustrates a case in which all LLMs and most human participants produced incorrect responses. The task required describing the relationship between CPS values and the relative abundance of cell types across two regions. In this instance, o1 grouped the two regions together and emphasized shared patterns, whereas the correct human responses separated analyses by both region and cell type. This outcome suggests that LLMs often overlook fine-grained details and require improved capacity for figure-specific inference. Human errors in this example primarily stemmed from not stratifying the analysis by cell type, highlighting the intrinsic difficulty of extracting precise information from scientific figures. A contrasting example is shown in Figure \ref{fig: lmm human evaluation metric} (e), where all LLMs failed but nearly all human participants answered correctly. The question asked participants to identify the method with the greatest variation across metrics. While LLMs incorrectly selected another option, their reasoning process appeared consistent with human logic, as both relied on interpreting boxplot widths. The discrepancy suggests that although LLMs followed the correct reasoning path, they failed in execution, whereas humans directly identified the correct method. All identified error categories are summarized in Supplementary File 1.

We also analyzed the error sources for both LMMs and human participants, categorizing them into three classes: (1) reasoning process errors, (2) information extraction errors, and (3) generation errors. As shown in Extended Data Figures \ref{supfig:error type analysis} (a) and (b), type 2 errors accounted for the largest proportion in both groups, while type 1 errors occurred less frequently. \textcolor{red}{Notably, human participants did not exhibit type 3 errors, underscoring their stronger ability to follow instructions (the number of type 3 errors is 0).} By contrast, smaller-scale LMMs occasionally produced generation errors, such as missing outputs or out-of-context responses. To further assess the alignment between humans and models, we compared the outputs of o1 with those of the best-performing human participant by projecting them into an embedding space and calculating cosine similarity (Extended Data Figure \ref{supfig:sim info} (b)). The results indicate residual differences between human- and LMM-generated outputs, suggesting an incomplete overlap in their interpretive processes. This observation motivates the exploration of reasoning-enhancement strategies, leveraging o1-type models to support human interpretation of scientific figures. Overall, these findings highlight both the alignment and divergence between LMMs and humans in addressing complex figure-based questions, offering valuable insights for advancing LMM figure comprehension through human feedback and, conversely, for augmenting human analysis with model-assisted reasoning.

\textcolor{red}{We also considered the analysis of cost for validating the correctness of answers by reporting the error rates from both LMMs and human experts, shown in Extended Data Figure \ref{supfig:error rate analysis}. According to our results, advanced LMMs such as o1 have a low error rate, and thus, their cost of verification is also the lowest among all LMMs. Open-source LMMs as well as weak LMMs produce more errors, and thus they have a higher cost. On average, the cost of verification for human outcomes is lower than the cost of verification for AI outcomes.}

\begin{figure}
    \centering
    \includegraphics[width=1\linewidth]{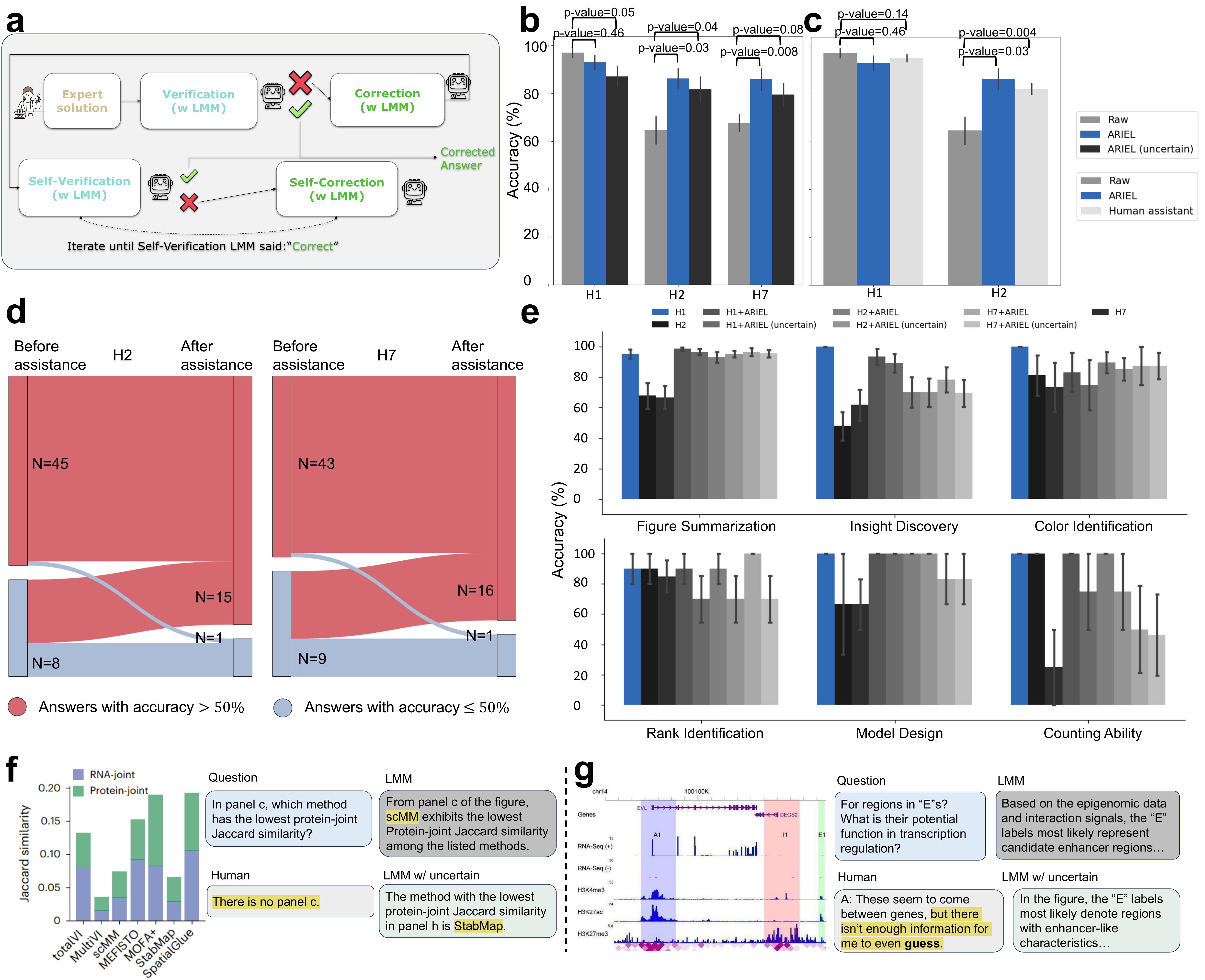}
    \caption{Results of verification-correction LMMs as collaborators for helping human researchers. We report the average and standard error in all bar plots. (a) Landscape of proposed self-correction pipeline by \method{}. The solid arrows in the panel indicate the steps that must be executed, and the dashed arrows indicate the steps that are optional. (b) Bar plots of comparisons of performances among human participators and human participators with \method{} (self-corrected LMMs) under two different modes. (c) Bar plots of comparisons of performances among human participators, human assistants, and \method{}. We averaged the scores of six human assistants to compare. (d) Sankey plots for demonstrating the improvement of using \method{} as a research assistant for human experts H2 and H7. We count the number of questions based on the conditions of with/without assistance and correctness by setting 0.5 as the accuracy threshold. (e) Bar plots of accuracy and standard errors from different options by problem categories. (f) One case study with wrong solutions coming from all options. (g) One case study with wrong solutions from humans, which is later corrected by LMMs.}
    \label{fig: correct answer}
\end{figure}

\textbf{Verification and Correction LMMs help humans understand scientific figures.} In this section, we demonstrate that LMMs can be effectively employed for verification and correction, thereby improving human interpretation of scientific figures in both qualitative and quantitative terms. The overall pipeline is summarized in Figure~\ref{fig: correct answer} (a), with subsequent panels presenting the results. Figure~\ref{fig: correct answer} (b) shows the accuracy of three participators (labeled H1, H2, and H7) under three conditions: the solution written by themselves without LLM assistance (pink bar), the solution is verified and corrected by an LLM assistant using our verification-correct framework (yellow bar for the default mode and green bar for the uncertain mode). Accuracy (ranging from about 40.0 to 100.0) is plotted on the vertical axis, while each participator’s results are grouped along the horizontal axis. Comparing the Raw (pink) settings, H1’s baseline accuracy is clearly higher than that of H2 and H7. When the LLM assistant (yellow) or LLM assistant uncertain (green) is applied, the corrected results of H2 and H7 (they originally performed at a lower accuracy) show measurable improvements. Statistically significant p‐values shown in the same figure indicate that these gains are meaningful. Overall, our results illustrate that human participators with lower initial analytical performance benefited notably from LMM assistance, improving their answer accuracy by a large margin. This observation is further supported by comparing the performances between \method{} and human assistants for correcting the answers from participators, shown in Figure \ref{fig: correct answer} (c) and Extended Data Figure \ref{supfig:human correct details}, which demonstrates that \method{} has a generally better performance compared with the average level of human experts. Meanwhile, we show the improvement of using \method{} as a corrector versus the raw answers via Sankey plots for H2 and H7. Based on Figure \ref{fig: correct answer} (d), \method{} can improve the accuracy of answers for over 65\% questions with low initial accuracy, further supporting the strong capacity of \method{} as a research assistant. Extended Data Figure \ref{supfig:lmm include human correct details} shows our consideration of inviting human evaluators using \method{} to correct the previous answers, and the final accuracy is similar to the outputs from \method{}, and thus we have demonstrated the potentials of using \method{} for biomedical research. 

Figure~\ref{fig: correct answer} (e) shows a breakdown analysis over different categories of biomedical questions. Applying an LLM assistant over H2 or H7 consistently outperformed H2 and H7 for all categories, indicating the effectiveness of the LLM assistant for verifying and correcting H2 and H7 solutions. Applying an LLM assistant in the uncertainty mode where ``I am not sure about the answer" is added after the human answer to deliberately encourage LLMs to correct the answer outperforms the original H2 and H7 performance in most categories except for rank identification. Our LMM output inspection shows that LLM generates relevant context, but it struggles to consistently produce a final ranked answer for the rank identification task when the uncertainty mode is enabled. Among all categories, Insight Discovery exhibited the largest variance in accuracy across both human and LLM-assisted answers, consistent with earlier findings. This result highlights that generating novel biological insights remains a challenge, reflecting limitations not only in current LMMs but also in human expertise.

We present case studies in Figures~\ref{fig: correct answer} (f) and (g) to further illustrate the effectiveness of the verification–correction framework. In Figure~\ref{fig: correct answer} (f), the human participant failed to identify the target panel, and the LMM assistant also produced an incorrect response. However, unlike the human, the LMM provided explicit reasoning, demonstrating a more systematic attempt at problem solving. In Figure~\ref{fig: correct answer} (g), the human participant was unable to interpret the figure due to its complexity, whereas the LMM successfully retrieved relevant domain knowledge, identified and extracted the patterns in the ``E" regions, and produced a correct explanation. Notably, human participants were permitted to consult external resources for this task, suggesting that the LMM exhibited a stronger ability to integrate genomic annotation knowledge in reaching the correct conclusion. These examples highlight two important points: (1) the LMM can correct human errors arising from knowledge gaps, and (2) the proposed verification–correction pipeline effectively elevates the performance of less analytically skilled participants to a level comparable with top human experts.

\begin{figure}
    \centering
    \includegraphics[width=0.9\linewidth]{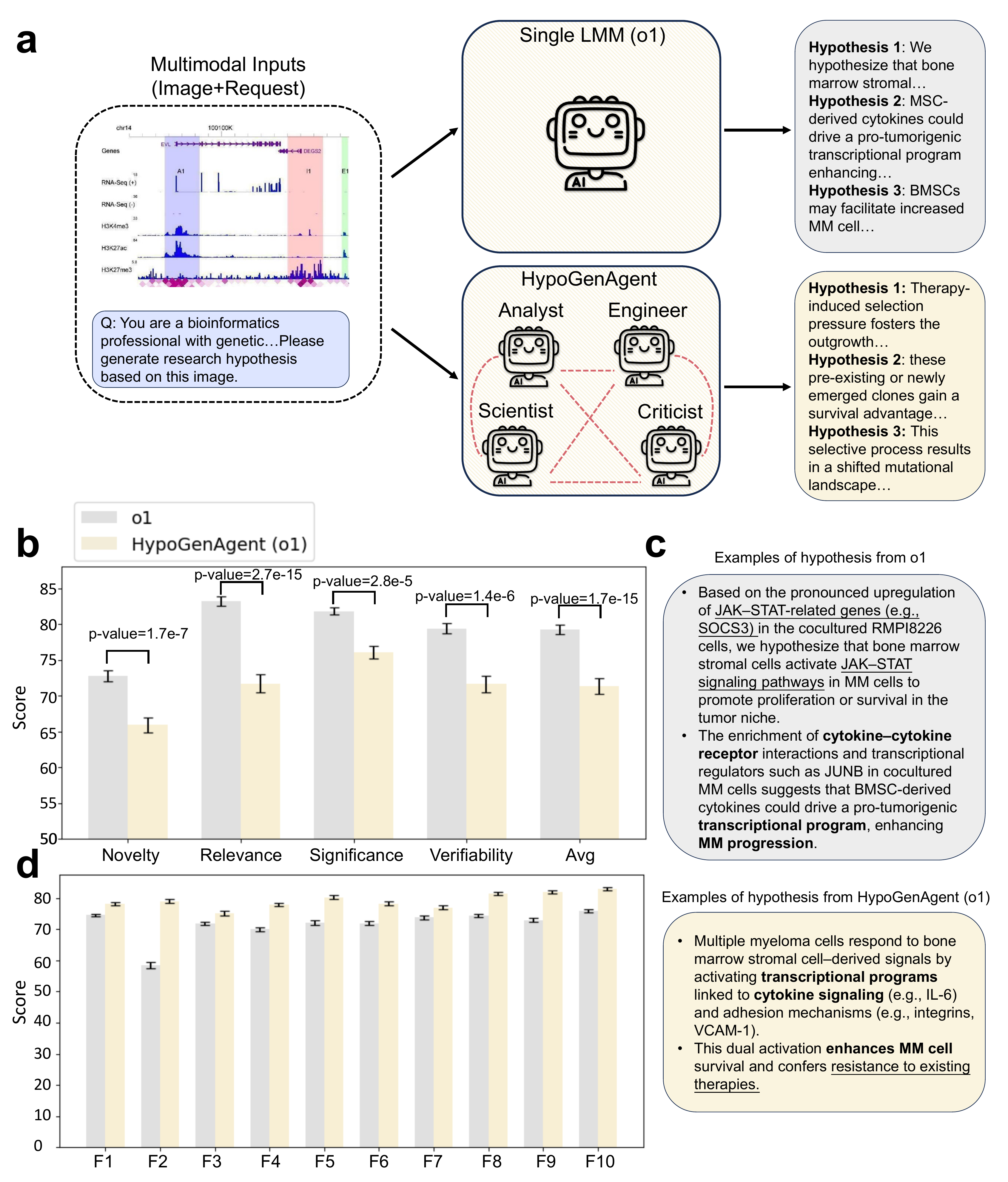}
    \caption{Landscape and results of generating hypotheses from multimodal inputs. We report the average and standard error in all bar plots. (a) Our proposed framework of generating scientific hypotheses with two different modes, which all generate hypotheses. The HypoGenAgent mode has four different LMMs as group members, and their functions are explained in the main text. (b) Bar plot of evaluation results of different generated outputs made by human experts with four metrics and their averaged score (Avg). (c) Examples of generated hypotheses with the same image inputs but from different modes. Important pathway and cellular activity information is boldfaced. The distinct information from different modes is underlined. (d) Evaluation results across different figures.}
    \label{fig: mmhypo gen}
\end{figure}

\textbf{Exploring the hypothesis-generation capacity of \method{} framework with text-image paired inputs.} Finally, we explore the potential of leveraging \method{} to integrate multimodal understanding for scientific hypothesis generation, with the goal of inspiring novel directions in biomedical research. In this analysis, we reuse scientific images from the previous tasks but apply new prompts specifically designed for hypothesis generation. We investigate two modes: (1) a single-LLM mode using o1 as the base model, and (2) a Multi-LMM agent mode (HypoGenAgent), also based on o1. HypoGenAgent employs four distinct LMMs assigned to different roles, following a modified design from \cite{qi2024large}. Here the Analyst takes the multimodal inputs and performs quantitative analysis, the Engineer executes the experimental plan, the Scientist proposes new hypothesis, and the Criticist can verify the quality of generated hypotheses. An overview of the framework is provided in Figure \ref{fig: mmhypo gen} (a). To evaluate the quality of the hypothesis generated, we consider different aspects modified from \cite{zhong2023goal} with a comprehensive framework to evaluate scientific insight.

\begin{itemize}
    \item Novelty: Does the hypothesis introduce new information or perspectives?

    \item Relevance: How closely is the hypothesis related to the topic or question?

    \item Significance: Does the hypothesis have am impact on understanding or addressing the problem?

    \item Verifiability: Can the hypothesis be tested using existing methods or data?
\end{itemize}

\textcolor{red}{Each evaluation criterion was quantified on a scale from 0 to 100, where 0 indicates complete non-compliance and 100 indicates full compliance. To mitigate potential bias in the evaluation process, we recruited 10 human experts, as well as Claude-Sonnet-3.7, as an automated judge \cite{chen2024mllm}. The scores provided by human experts can be found in Figures \ref{fig: mmhypo gen} (b) and (d). According to these figures, o1 always has a higher metric-specific score compared with HypoGenAgent (o1). Moreover, we perform statistical testing to compare the results from two methods, and the significance further shows that o1 works generally better than HypoGenAgent (o1) in hypothesis generation ($\text{p-value}<0.05$ with two-sided Mann-Whitney U test across 10 images). This is different from the evaluation made by the AI judge (Extended Data Figure \ref{supfig: llmasajudge_image}), and thus, human experts are more reliable evaluation sources, and a multi-agent system cannot directly help in proposing better hypotheses for discovery.}

\textcolor{red}{To better understand the properties of the hypotheses generated by these modes, we select representative examples of generated hypotheses for in-depth analysis. Based on Figure \ref{fig: mmhypo gen} (c), we found that the hypothesis from o1 generally has a longer length and more figure-oriented information, while HypoGenAgent (o1) tends to generate hypotheses with shorter lengths and more general plans. Moreover, o1 tends to discuss more gene-function-level information, while HypoGenAgent (o1)'s outputs focus more on therapeutic discovery. Therefore, there exists a completeness-conciseness balance when using \method{} as a hypothesis generator, which may help researchers to design ideas that fit well with their requirements.}

\textcolor{red}{We have also considered Fleiss’ kappa (FK) \cite{fleiss1971measuring,virtanen2020scipy} scores across human experts in four different experiments (text summarization, figure understanding, idea novelty evaluation, and summarization quality evaluation) to examine the consistency of metrics. According to Extended Data Figures \ref{supfig: consistency check} (a)-(d), we showed the heatmap across participants for each sample and computed the FK score, and we found that there are no obvious patterns in all tasks, and the FK score is also low (nearly 0). Therefore, we did not figure out obvious similarity across human evaluators, and thus our evaluation outcomes admit the differences across human experts and also cover diverse conditions. We reported the demographic information (nationality, sex, educational background, etc.) of selected human experts in Supplementary File 2, which could partially explain the source of such divergence.}

%% file: section_folder/Discussion.tex
Investigating and analyzing the ability of foundation models to solve scientific problems can provide meaningful summarization for the current stage and an invaluable guide for the future development of these models. Therefore, in this manuscript, we present a novel framework, named \method{}, which can evaluate the capacity of current advanced LLMs and LMMs in summarizing long research papers and understanding complicated scientific figures. We not only reported the performances of different models in these tasks but recruited human participators with expert-level biomedical knowledge and compared the human outputs with model outputs to further investigate the performance difference and preference difference. Finally, we demonstrated that applying specific strategies can enhance the ability of models or human participators in addressing these two tasks, and thus \method{} can also work as an interaction platform between human researchers and foundation models for understanding research background, defining scientific question, and accelerating scientific discovery.

For the long document summarization task, we found that LLMs without fine-tuning generally could not perform well or generate meaningful abstracts, which might be caused by the limitation of context window size and the shallow understanding of domain-specific knowledge. Therefore, we considered adjusting the prompts (prompt engineering) and fine-tuning the base model (fine-tuning) to evaluate the updated results. While prompt engineering did not generate results with much difference, fine-tuning an LLM with domain-specific knowledge can enhance the model's capacity to summarize these research manuscripts. Our experimental results suggest that introducing more biomedical knowledge can help us construct a better scientific text reader and summarizer.

When we considered the performances of human participators in summarizing long documents, we found that they generally did not provide outputs with high scores by comparing them with answers from the fine-tuned LLMs. Moreover, humans also demonstrated variation in the length of generated abstracts and higher costs in reading and summarizing scientific figures, which further supports the necessity of including LLMs as helpers in this task.

For the scientific figure understanding task, we found that base LMMs without test-time computation and reasoning ability cannot make precise answers based on the questions and images as inputs. Models like o1 can achieve higher scores than other LMMs, but can only outperform two human participators in the general comparison. After analyzing the differences between the outputs from models and humans, we developed a new pipeline to improve the answers from humans with self-verification and self-correction processes based on o1. Our experimental results showed that this approach could significantly improve certain human participators' performances, which reveals the potential of using the foundation model as an assistant for understanding complicated multi-modal problems. Considering the high costs of human participation, including foundation models in the loop of scientific research can also provide a more efficient solution for saving human resources.

Furthermore, our results based on difficulty analysis imply that LLMs even have consistency with humans in their understanding of difficult problems. For example, LLMs also usually summarized abstract with low scores on the articles that most human participators marked as most difficult. Meanwhile, the answers for questions with the insight discovery category in the scientific figure understanding task also had larger variation compared with other categories, which was observed in both outputs from humans and outputs from models. These findings provide a new perspective on generation errors in foundation models. Finally, we demonstrated that \method{} can also work as a hypothesis generator with multimodal inputs for accelerating scientific discovery, which aligns well with the perspectives of having agentic AI for accelerating scientific research \cite{koutra2025towards}.

Despite these findings from our research, there are several limitations in our experiment. First, the technology for training the foundation models has been advancing, especially for the reasoning capacity, so stronger models are likely to emerge in the near future. Second, although we have tried various approaches to recruit as many human participators as possible, due to resource and motivation constraints, we were unable to recruit more qualified participators, and thus our conclusions might be constrained by our sample size. Finally, our design of datasets could be more refined and fine-grained. Although we tried to control the focus on biomedical research, we could potentially give more detailed optimizations for specific problems if we had access to more subcategories.

\textcolor{red}{To conclude our research, we summarize the following key discoveries based on the experimental results involving FMs and human participators: 1. For the long document summarization task, current LLMs with zero-shot inference mode cannot provide high-quality summaries, while fine-tuning an LLM can improve its capacity for this task. Moreover, outputs from humans generally have lower scores compared with outputs from the fine-tuned LLM. Fine-tuned LLM can provide more precise summaries for papers with less numerical information. 2. For the scientific figure understanding task, LMMs without reasoning improvement cannot achieve high accuracy across all categories, while large multi-modal reasoning models such as o1 can make improvements in understanding complicated figures. Moreover, o1 cannot exceed most of the human participators, demonstrating the challenges in the real applications of these models. 3. With the help of verification and correction strategy, we can leverage the reasoning ability of advanced LMMs to help under-performing human participators by correcting their wrong answers, and thus improve their performances. Therefore, our experiment shows the potential of using LMMs as assistants for understanding complicated scientific figures. There is a great potential of using foundation models as assistants to reduce the workload and cost in biomedical research, which can facilitate the advancement of this field.}


%% file: section_folder/Methods.tex
\textbf{Problem Statement.} In this study, we focus on evaluating the outputs of LLMs on biomedical manuscript understanding and comparing abstracts or answers generated by humans and LLMs. Here we define the participator $\mathcal{M}$ in the experiment as $O=\mathcal{M}(P,T,I)$, where $O$ represents outputs, $P$ represents prompts, $T$ represents text inputs, and $I$ represents image inputs. To evaluate the quality of LLMs-generated abstracts, $I$ is placed as empty, and we treat the original abstracts as ground truth $O^T_G$ and compare their differences by both quantitative approaches (denoted as evaluator $E(O_G,O)$) and qualitative approaches. We also invite human experts to summarize the same testing documents and compare the differences in outputs. To evaluate the quality of LMM-generated answers for figure understanding, $I$ is paired with question $P$, and $T$ is empty. Here we compare their outputs with human-annotated ground truth $O^I_G$ and invite human experts to answer the same questions and evaluate both human- and LMMs-generated outputs.

\textbf{Datasets in \method{}.} In this study, we propose a new dataset designed for evaluating the ability of FMs for long document summarization and scientific figure understanding, and thus our dataset contains two parts. The first part is designed with instructions, input texts, and ground truth. The second part is designed with instructions, input texts, input images, and ground truth.

Regarding the first part of the evaluation dataset, we download the open-access research manuscripts from PubMed \cite{white2020pubmed} with a knowledge cut-off date later than 2024 January 1st and earlier than 2024 June 1st. The total dataset contains $n=2571$ papers, covering different topics. We further split the papers into training and testing datasets with a ratio of 0.7/0.3. The training dataset was used to fine-tune specific models, while the testing dataset was used to evaluate both open-source and closed-source models. For one paper, we took the abstract section out as ground truth, and deleted this section in the original paper as input texts. Our instruction was used to query the model to give us the summarization of targeted research manuscripts. Our evaluation does not suffer from the problems of data leakage, as our setting is harder and different from the generative pre-training stage. By removing the abstract from the context, tested LLMs cannot know directly infer the abstract based on the causality in text.

Regarding the second part of the evaluation dataset, we collected 20 scientific figures from published manuscripts in top-tier journals focusing on genomics and proteomics research \cite{wang2024scientific, long2024deciphering, dini2024multiplex}. We designed several questions to evaluate the performances of LMMs with different aspects, including figure summarization, color identification, model design, rank identification, recommendation, and insight discovery. We adjusted the prompt to provide necessary background information with questions as instructions and input texts, and combined the prompt with the paired figure as inputs of different LMMs. We generated the ground truth information of each question by referring to the original description of these figures in the manuscript. Our evaluation does not suffer from the problems of data leakage, as the questions are designed by us without using AI models, and our selected figures have copyrights.

\textbf{Automatic evaluation framework.} After obtaining the generated outputs of different FMs, we collected the ground truth information paired with model outputs and performed a quantitative assessment with various metrics.

For the long document summarization task, we consider seven different metrics for evaluation, including BLEU, ROUGE-1, ROUGE-2, ROUGE-L, BERT score (BERT), F1-Radgraph, and MEDCON, for evaluation \cite{papineni2002bleu, lin2004rouge, zhangbertscore, jain1radgraph, yim2023aci}. Details of the metrics are introduced below:

1. BLEU: The BiLingual Evaluation Understudy (BLEU) score measures the quality of generated texts by extracting the grams of generated texts and ground truth texts and then comparing the similarity based on two sets of grams. This score ranges from 0 to 1 and is later scaled to (0,100). A higher score means better model performance.

2. ROUGE: Recall-Oriented Understudy for Gisting Evaluation (ROUGE) score measures the F1 score based on the n-grams from the generated texts and ground truth texts. It treats the grams in the generated texts as prediction and ground truth texts as labels, and uses the length of selected texts and the number of grams to compute precision, recall as well as F1 score. ROUGE-1 represents using 1-gram, ROUGE-2 represents using 2-grams, and  ROUGE-L represents using the longest common subsequence. This score ranges from 0 to 1 and is later scaled to (0,100). A higher score means better model performance.

3. BERT: Bidirectional encoder representations from transformers (BERT) model is pre-trained with large-scale texts for language understanding, which also has a strong ability in generating the representations of texts. Therefore, BERT score utilizes the similarity of embeddings from generated texts and ground truth texts based on BERT. This score ranges from 0 to 1 and is later scaled to (0,100). A higher score means better model performance.

4. F1-Radgraph: Radgraph is a method for extracting clinical entities and relations from radiology reports. We use Radgraph to extract such information from generated texts and ground-truth texts and compute the F1 score. This score ranges from 0 to 1 and is later scaled to (0,100). A higher score means better model performance.

5. MEDCON: MEDCON restricts the concepts and entities in QuickUMLS \cite{soldaini2016quickumls} semantic groups (Anatomy, Chemicals \& Drugs, Device, Disorders, Genes \& Molecular Sequences, Phenomena and Physiology) and we extract such sets from generated texts and ground truth texts. We then compute the F1 score based on these two sets. This score ranges from 0 to 1 and is later scaled to (0,100). A higher score means better model performance.

Based on the evaluation results recorded in \cite{van2024adapted}, BERT and MEDCON scores have a high correlation with the correctness of outputs, while BLEU and ROUGE scores have a high correlation with the completeness of outputs. We compute the final score of each method by averaging the metric scores across all the testing manuscripts. The scores are scaled all into 0-100 and higher scores mean better model performances.

For the scientific figure understanding task, by following the design of \cite{xiao2024logicvista, wang2024scientific}, we only consider the accuracy by comparing model outputs and ground truth information. Since we also need to consider the inference information as well as format information of model outputs, and thus our previous settings for long document summarization evaluation cannot be directly applied. We utilize Spacy \cite{Honnibal_spaCy_Industrial-strength_Natural_2020} as a tool to split model outputs into different sentence blocks with similar methods and evaluate the correctness of each block. The accuracy is computed based on the number of correct blocks and the total number of blocks, for each question. For each figure, we compute the average score across different questions to report the performance of specific models in the given figure.

{\color{red}
\textbf{Human-Eval framework.} We invited experts with strong biology backgrounds to evaluate the model outputs from these tasks. Each expert was guaranteed to have doctoral-level knowledge in this domain, and they did not know other participators' identities; thus, it is hard to cheat in this experiment. We have several stages for each task to make a comprehensive evaluation. We ensured that the human interviewers and LLMs or LMMs had the same inputs.

For the long document summarization task, we assigned ten papers to every expert, and the expert would read the manuscript without the information of the abstract and write down the abstract in 100-200 words. We then compared the human outputs with LLM outputs to determine which one is better by measuring the similarity between outputs and observed abstracts. 

For the scientific figure understanding task, the whole process is similar. The experts would read the image and write down the answers to the same questions used in evaluating LMMs. We also requested the experts to provide the uncertainty level of their answers. At this stage, experts were not allowed to use LMMs as assistants. Moreover, we asked experts to correct other participators' answers based on their knowledge, and we allowed them to use LMMs to assist them in this stage. We collected human outputs in different stages and reported accuracies based on the results.

For the summary quality evaluation task, we assigned ten papers paired with summaries from three sources (original paper, best LLM, and one human expert) and invited different human experts to evaluate their quality from five perspectives:

\begin{itemize}
    \item Completeness: Definition: Measures whether the output fully addresses all required aspects of the task or question. Evaluation Criteria: Covers all key components requested; Does not omit critical steps, assumptions, or conclusions; Includes necessary context, explanations, or examples (when required).
    
    \item Relevance: Relevance evaluates how well the model output aligns with the user’s intent and stays focused on the given task. A high score reflects content that directly responds to the prompt and avoids unnecessary or unrelated information. Responses that include tangential, off-topic, or distracting material should receive lower scores, especially if such content detracts from addressing the core request.
    
    \item Conciseness: Conciseness assesses how efficiently the model communicates information without unnecessary verbosity. A high score indicates that the response is succinct, avoids redundancy, and includes only the level of detail appropriate for the task. Scores should be reduced for outputs that are overly long, repetitive, or padded with filler content, particularly when verbosity obscures the main message.
    
    \item Coherence: Coherence measures the logical organization and flow of the response. High-scoring outputs present ideas in a clear, well-structured manner, with smooth transitions and consistent reasoning throughout. Lower scores should be given when the response is disorganized, contains abrupt shifts, internal contradictions, or a progression of ideas that is difficult to follow.
    
    \item Clarity: Clarity evaluates how easily the response can be understood by the intended audience. A high score reflects precise language, well-formed sentences, and unambiguous explanations, with technical terms defined when necessary. Responses that are vague, confusing, or difficult to interpret should receive lower scores, especially if lack of clarity hinders comprehension of the main points.

\end{itemize}

The score of each perspective is in the range between 0 and 100. We ensure that human experts in this experiment will not be assigned summaries written by themselves.

For the idea quality evaluation task, we assigned ideas generated by LMMs and AI agents from ten figures to different human experts and invited them to evaluate their quality from four perspectives:

\begin{itemize}
    \item Novelty: Does the hypothesis introduce new information or perspectives?
\item Relevance: How closely is the hypothesis related to the topic or question?
\item Significance: Does the hypothesis have impact on understanding or addressing the problem?
\item Verifiability: Can the hypothesis be tested using existing methods or data?

\end{itemize}

The score of each perspective is in the range between 0 and 100. 

Overall, we recruited 18 human experts in this experiment. Seven people were assigned with papers for the summarization task, eight people were assigned with questions and figures for the scientific figure understanding task, and six people were assigned with questions, figures, and answers from different resources for the human-verification-correction task. The questionnaire we used as instructions can be found in Supplementary File 3.}

\textbf{LMM verification and correction to improve figure understanding capacity of humans.}
Scaling LMM test-time compute has been shown to be effective in improving the reasoning performance of LMMs on math and logical reasoning\citep{chen2025setsleveragingselfverificationselfcorrection, snell2024scaling, sui2025metareasonerdynamicguidanceoptimized}. We adopt the self-verification and self-correction component of SETS \citep{chen2025setsleveragingselfverificationselfcorrection}, a test-time scaling framework for iterative verification and correction and propose a verify-correct framework using an LMM to conduct verification and correction over human-written solutions for biological figure reasoning problems. In contrast with SETS which adopt self-verification and self-correction over LLM outputs, we conduct verification and correction over human outputs as well as iterative self-verification and self-correction over LMM-generated outputs. 

We illustrate the entire process in panel a in Figure~\ref{fig: correct answer}. Specifically, in the first verification step, an LMM $\mathcal{M}$ verifies whether the human-written solution is correct. This is done with a specially crafted verification prompt $P_V$ that asks the model to break down the solution and then think step by step to determine if the solution is correct given the figure and question. A verification step is followed by a correction step. If the model deems a solution incorrect, it refines that solution by addressing the specific errors it identified in the verification step. The refined solution generated by the LMM will be fed back into the same LMM it will verify the correctness of the answer generated by itself. This will be followed by a self-correction step where the same LMM will correct the answer if it deems that it is incorrect in the self-verification step. The self-verification and self-correct steps will continue iteratively until a model deems a solution to be correct, where the refinement process will terminate, and the solution will be evaluated for correctness by an expert annotator. Prompts we used for verification and correction are in the Appendix \ref{append:prompts}. 

We use one of the most advanced general-purpose Multimodal reasoning models, OpenAI o1 \citep{jaech2024openai} for evaluating LLM verification and correction over improving figure understanding capacity of humans. The temperature is set to 1 to promote the model to generate diverse and creative reasoning steps. The maximum iterations of verification and correction are set to 10 by following the general setting in this method \citep{chen2025setsleveragingselfverificationselfcorrection} (In our task, observed minimal iteration is 6, and maximal iteration is 10). The workflow is illustrated in Algorithm 1.

\begin{algorithm}[htbp]
\caption{Self-Verification and Self-Correction pipeline.}
\begin{algorithmic}[1]
\Statex \textbf{Input:} System prompt $T_S$, question $Q_M$, scientific image $I_S$, human outputs $O_H$, verify prompt $T_V$, correct prompt $T_C$, number of iteration $N$.
\Statex \textbf{Helper Models:} An LMM $\mathcal{M}(\cdot,\cdot,\cdot)$, concatenation function $\cdot||\cdot$.
\Statex \textbf{Output:} Corrected outputs $O_C$
\State INIT: initialize all parameters.
\If{$\mathcal{M}(T_S||T_V,Q_M||O_H,I_S)$ is True}
\State $O_C=O_H$
\State Return $O_C$
\EndIf
\For{$i$ in $N$ steps}
\State $O_i=\mathcal{M}(T_S||T_C,Q_M||O_H,I_S)$
\If{$\mathcal{M}(T_S||T_V,Q_M||O_i,I_S)$ is True}
    \State $O_C=O_i$
    \State Return $O_C$
\Else
    \State $O_H = O_i$
\EndIf
\EndFor
\State $O_C=O_i$
\State Return $O_C$
\end{algorithmic}
\end{algorithm}

We propose two different modes for this process, including default mode and uncertain mode. The uncertain mode means we add ``I am not sure about the answer." at the end of each human output to let the LMM agent be aware of the uncertainty level of such answers.

\textbf{Hypothesis generation.} To generate hypotheses from multi-modal inputs, we consider two different modes. The zero-shot mode is based on querying o1 with image-question pairs. The agent mode is based on querying HypoGenAgent (with o1 as the base model) with the same pairs. HypoGenAgent contains four different LMMs with different system prompts, including Analyst (extracting the important information from the context of inputs), Engineer (search from the knowledge bases to propose initial plans), Scientist (formulate the known resources into scientific hypotheses), and Criticist (Assess the feasibility of assumptions). The four LMMs can collaborate with each other to formulate an agent system for hypothesis generation. Our input prompts used in this section are reported in Appendix \ref{append:prompts}.

\textbf{Baseline explanation.} We select baseline models with the consideration of avoiding potential testing dataset leakage. 

For the long document summarization task, we consider Llama 3.1 (8B) \cite{touvron2023llama}, Mistral (7B) \cite{jiang2023mistral}, Gemma1 (8B) \cite{team2024gemma}, ChatGLM4 (9B) \cite{glm2024chatglm}, ChatGLM\_lora, Gemini-Flash-1.5 \cite{team2023gemini}, Claude-Sonnet-3.5 \cite{antrop2024}, GPT-3.5 \cite{papergpt3}, GPT-4 \cite{achiam2023gpt}, GPT-4o \cite{hurst2024gpt}, GPT-OSS (20B) \cite{gptoss2025}, GPT-OSS (120B) \cite{gptoss2025}, and Deepseek-v3 \cite{liu2024deepseek} as baselines. We do not require models' ability of accepting images as inputs. The former five models are open-source LLMs, while the latter two models are closed-source FMs. Llama 3.1 (8B) is an open-source LLM proposed by Meta with diverse training datasets. Mistral (7B) is an open-source LLM proposed by Mistral. Gemma1 (8B) is an open-source LLM from Google. We use the We use the instruction-tuning mode of these three models. ChatGLM4 (9B) is an open-source LLM proposed by ZhipuAI, and ChatGLM\_lora is the fine-tuned version of this model using the training dataset from PubMed data. GPT-3.5, GPT-4, and GPT-4o are three closed-source models from OpenAI, and GPT-4 (as well as GPT-4o) is more advanced than GPT-3.5. GPT-OSS series are open-source models from OpenAI. Deepseek-v3 is an advanced 670B large language model.

For the scientific figure understanding task, we consider Qwen-VL-Chat \cite{bai2023qwen}, LLaVA-13B \cite{liu2023visual}, Gemini-Flash-1.5, GPT-4V, GPT-4o, o1, and Claude-Sonnet-3.5/3.7 for evaluation. The former two methods are open-source, while the latter four models are closed-source. Qwen-VL-Chat is a large-scale LMM proposed by Alibaba Cloud. LLaVA-13B is an open-source LMM proposed by researchers from the University of Wisconsin–Madison and Microsoft Research. Gemini-Flash-1.5 is a closed-source LMM proposed by Google. GPT-4V, GPT-4o, and o1 are three closed-source LMMs proposed by OpenAI. GPT-4o is a more advanced version than GPT-4V, while o1 provides stronger reasoning ability. We also consider the five major voting (o1 (5 mv)) baseline, which means we ask o1 five times and collect the outputs with the highest frequency as the final output; and best-of-5 oracle (o1 (bo5)), which means we determine if o1 can make correct answers among these five generated outputs. We also consider other reasoning models such as DeepSeek-R1 \cite{guo2025deepseek}, but it does not support complicated figures as inputs. Claude is a closed-source LMM proposed by Anthropic.

\textbf{Dataset Availability.} In order to ensure that the datasets are valid and will not be used for model training in the future, we will provide access to datasets in Huggingface after approval. Please visit our Huggingface website and sign up for related documents if you are interested in using it. We do not plan to release human outputs to protect the privacy of participators and mitigate the negative effect caused by this study for participators.

\textbf{Codes Availability.} Our codes can be found in: \url{https://github.com/HelloWorldLTY/ARIEL}, and the datasets are hosted on \url{https://huggingface.co/datasets/iLOVE2D/ARIEL_DATA} (to be released after peer review) under the MIT license. Experiments with open-source models were run on a single NVIDIA H100 GPU. To reproduce our experiment results with closed-source models, we use official API with random state 2024 and default parameters. 

\textbf{Institutional Review Board (IRB) Approval.} This project has received approval from Yale IRB, with project number 2000039055. 

\textbf{Position Statements} Here we disclose the authors’ anticipated outcomes of the study before the experiment was conducted to be transparent about experimenter biases. All authors tend to keep the null hypothesis, which means FMs and humans can be fair or FMs performed worse than humans.



\section{Acknowledgments.}
This project is partly funded by OpenAI research program, Google Cloud research program, as well as NIH grants U01 HG013840, U24 HG012108, and P50 CA196530.

\section{Author contributions.}
T.L. proposed the study. T.L., S.H., X.L., H.W. and P.L. designed the framework. T.L. and S.H. ran the experiments and performed computation analysis. Members in Human-Eval Group contributed to evaluation results as human outputs. All authors contributed to writing. H.Z. supervised this project.  

\section{Competing interests.}
The authors declare no competing interests.

%% file: section_folder/Appendix.tex
\counterwithin{figure}{section}
\renewcommand{\figurename}{Extended Data Fig.}
\renewcommand\thefigure{\arabic{figure}} 

\section{Prompts}
\label{append:prompts}
The default prompt we used to ask LLMs for text summarization is: Please summarize the following text...

The meta prompt for text summarization is: I'm here to assist with all your biology-related inquiries. Whether you're a high school student struggling with genetics, a college professor teaching molecular biology, or an enthusiast eager to discuss the latest in conservation biology, I'm equipped to support you. My expertise covers a wide range of biological sciences including cell biology, genetics and genomics. I provide accurate, up-to-date information, tailored study guides, and can help develop research questions or experimental designs. Ask me anything from basic concepts to complex theories, and I’ll provide detailed explanations and visual aids to foster your understanding. Let’s explore the wonders of biology together!

The COT prompt for text summarization is adding ``Let's think it step by step." at the end of each input.

The prompts we used to ask LMMs for scientific figure understanding (including meta prompts and questions) are summarized in Supplementary File 4.

The prompts we used for self-verification and self-correction with LMMs are shown in Table~\ref{tab:prompts-verification-correction}.

\begin{table*}[h]
\centering
\setlength{\tabcolsep}{3pt}
\small
\begin{tabular}{p{15.5cm}}
\toprule
\textbf{Verification Prompt} \\ 
You are given a QUESTION and a PROPOSED SOLUTION. Your job is to: \\ 
1. Break down each component of the proposed solution. \\ 
2. Think step by step to verify if the proposed solution is correct given the question and the figure. \\
3. Write a line of the form ``The proposed solution is correct" or ``The proposed solution is incorrect" at the end of your response based on your analysis. \\ 
QUESTION: \{question\}. \\ 
PROPOSED SOLUTION: \{solution\} \\ 
\midrule
\textbf{Correction Prompt}
You are also given a question and a solution for the question. Your job is to outline your step-by-step thought process for deriving a new solution and also write down the new solution. \\ 
Question: \{question\} \\ 
Solution: \{solution and analysis\} \\
\bottomrule
\end{tabular}
\caption{Prompts used for verification and correction of human-written solutions and LMM-generated solutions.}
\label{tab:prompts-verification-correction}
\end{table*}

The prompt we used for hypothesis generation is:

\begin{quote}
\textbf{Research Background:} \{\texttt{background}\}

\medskip

\textbf{Hypothesis:} \{\texttt{hypothesis}\}

\medskip

\textbf{Objective:} Please score the hypothesis with this question: \{\texttt{select\_question}\} 0 is absolutely no and 100 is absolutely yes. Only provide the score as a number. For example: 10.
\end{quote}

\clearpage

\section{Supplementary Figures}

\begin{figure}[H]
    \centering
    \includegraphics[width=1\linewidth]{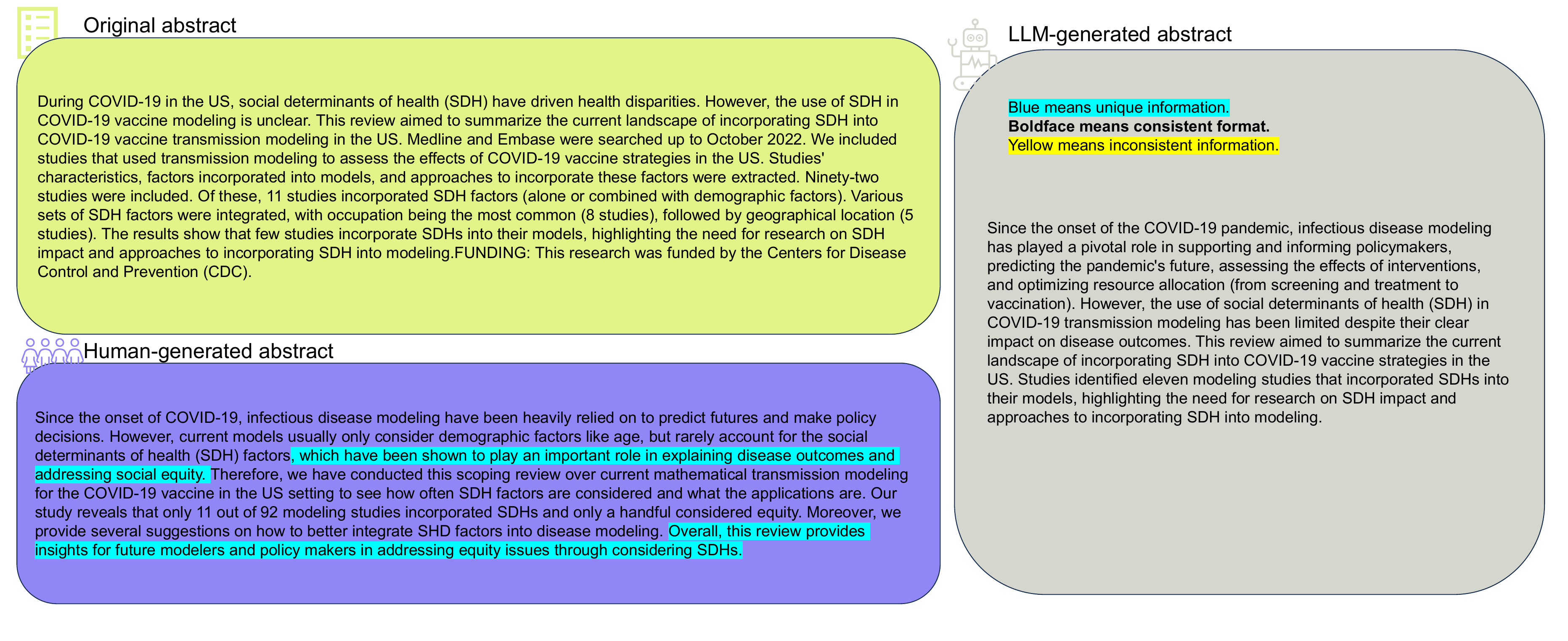}
    \caption{An example of human's and LLM's answers for text summarization.}
    \label{supfig: text enrich example}
\end{figure}

\clearpage
\begin{figure}
    \centering
    \includegraphics[width=1\linewidth]{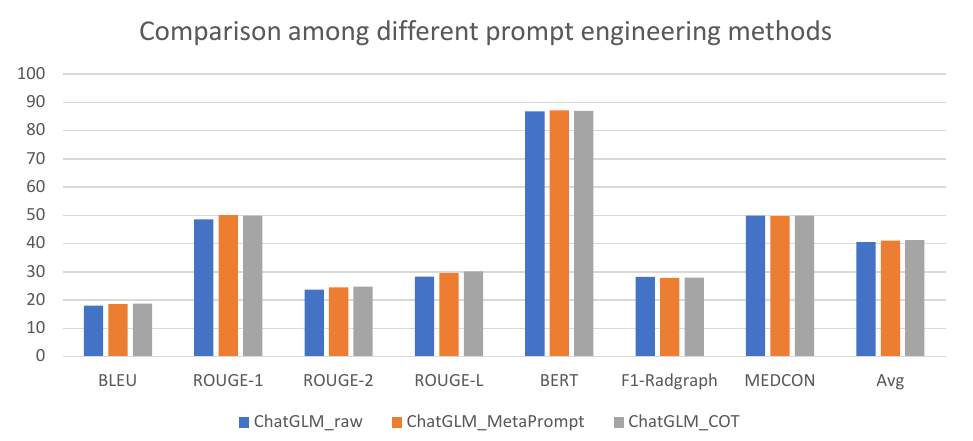}
    \caption{Results of prompt engineering for the text summarization task.}
    \label{supfig: prompt eng text task}
\end{figure}

\clearpage
\begin{figure}
    \centering
    \includegraphics[width=1\linewidth]{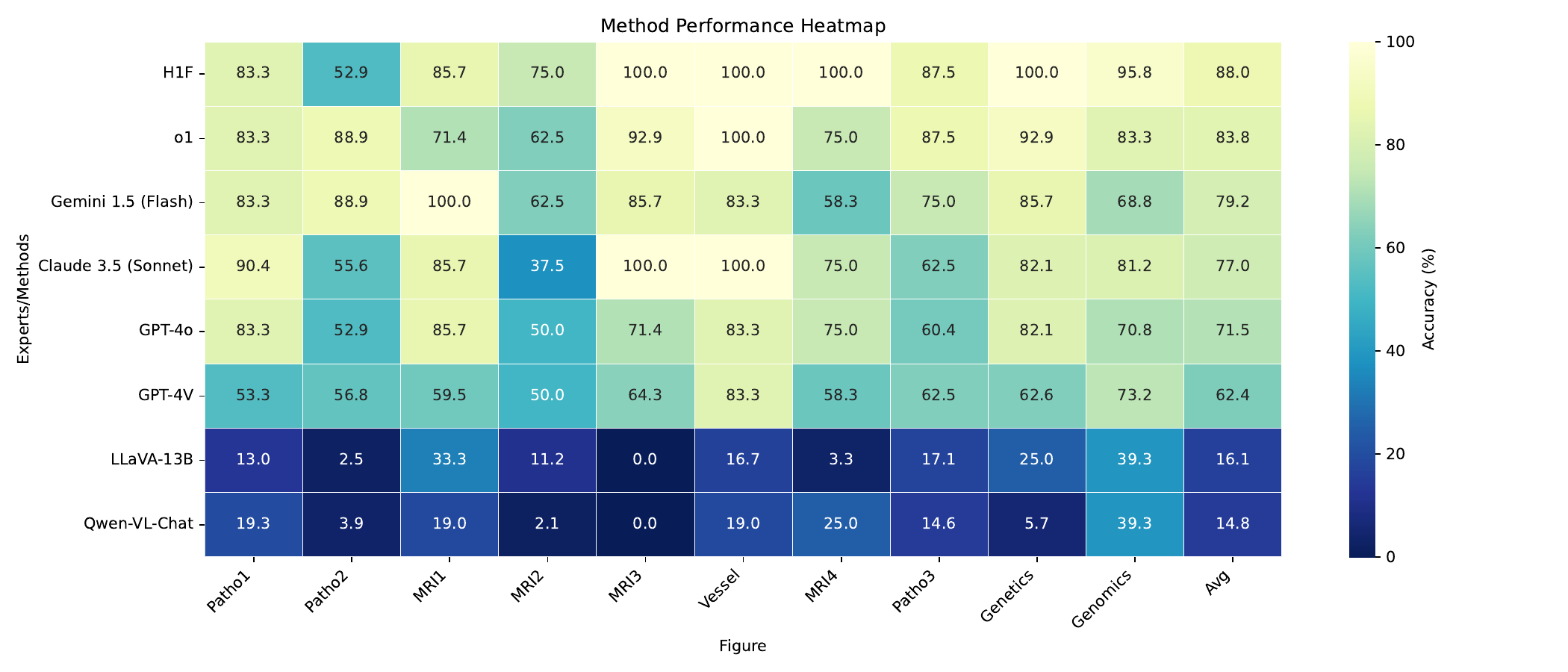}
    \caption{Performances of human experts and AI models for the figure understanding task, based on the additional images.}
    \label{supfig: additional image}
\end{figure}

\clearpage
\begin{figure}
    \centering
    \includegraphics[width=1\linewidth]{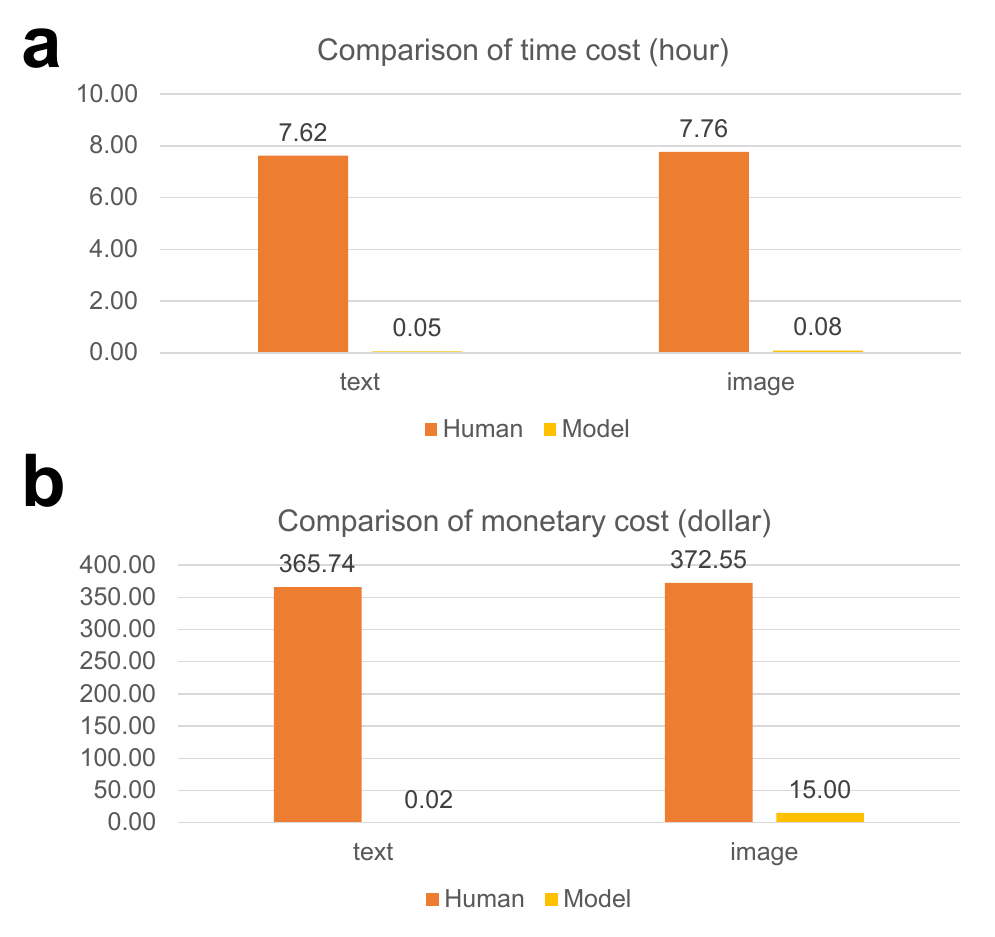}
    \caption{Comparison of time cost and monetary cost between humans and models. (a) Comparison based on the time cost across different tasks. (b) Comparison based on the monetary cost across different tasks.}
    \label{supfig: time cost compare}
\end{figure}

\clearpage
\begin{figure}
    \centering
    \includegraphics[width=1\linewidth]{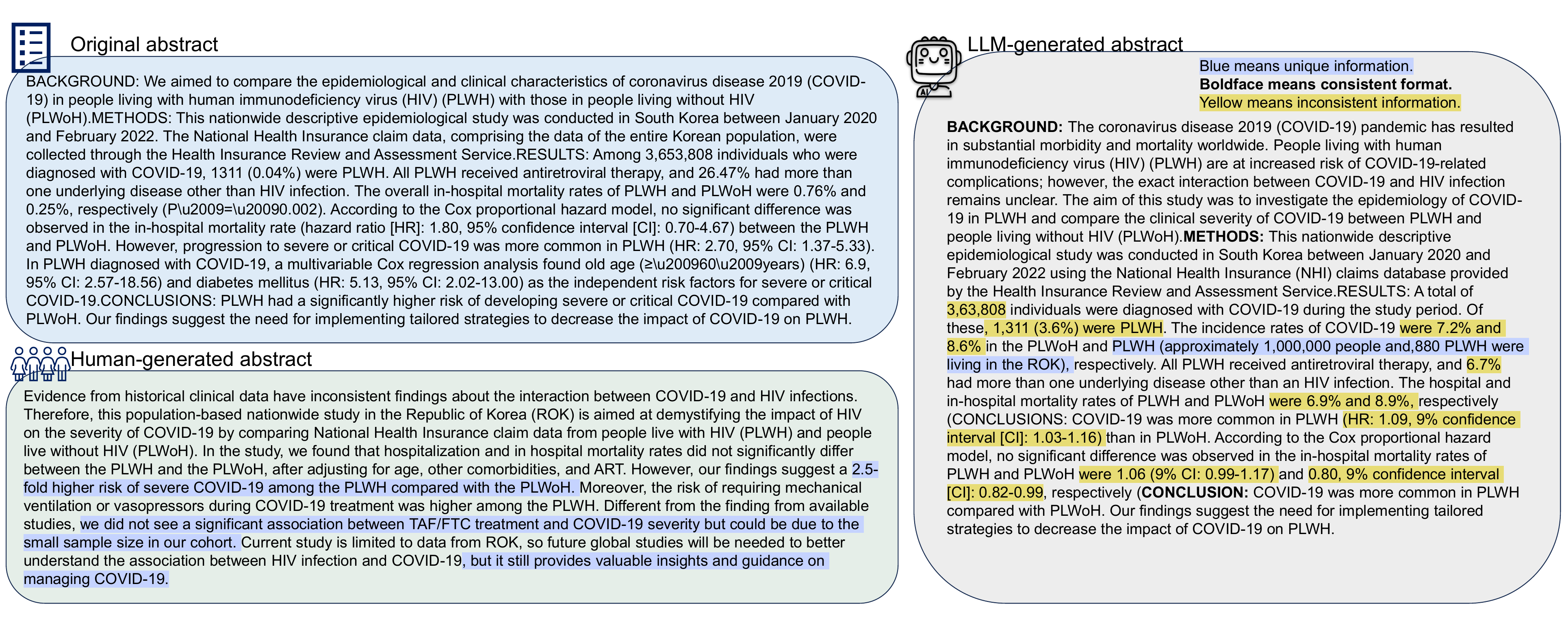}
    \caption{Case study based on the original abstract, human-generated abstract, and LLM-generated abstract.}
    \label{supfig: text sup info}
\end{figure}

\clearpage
\begin{figure}
    \centering
    \includegraphics[width=1\linewidth]{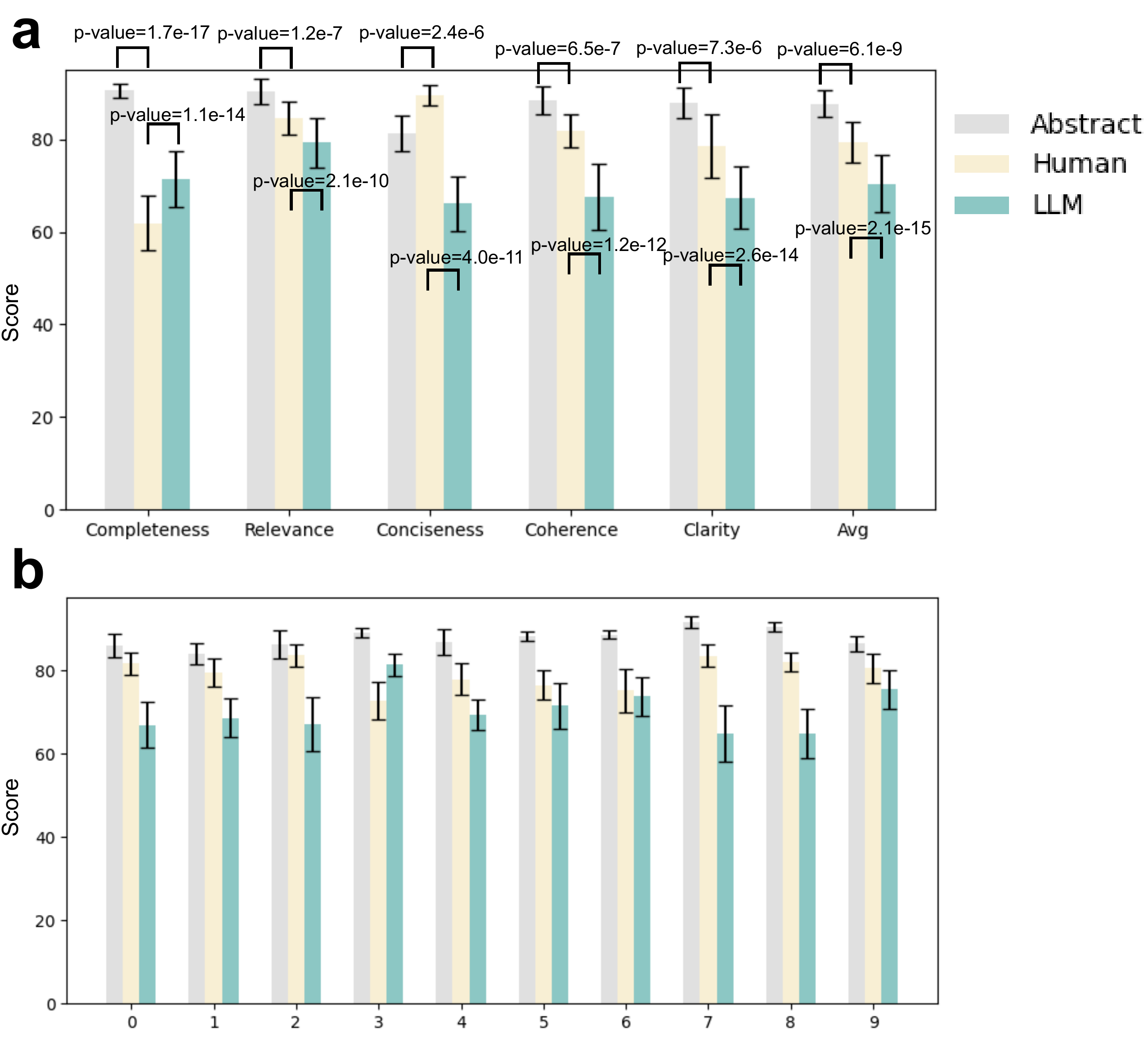}
    \caption{Humane evaluations for the quality of abstracts from three different resources (original abstract, human-generated abstract, and LLM-generated abstract). (a) Evaluation scores by metrics. (b) Evaluation scores by manuscript id.}
    \label{supfig: llm generated score}
\end{figure}

\clearpage
\begin{figure}
    \centering
    \includegraphics[width=0.6\linewidth]{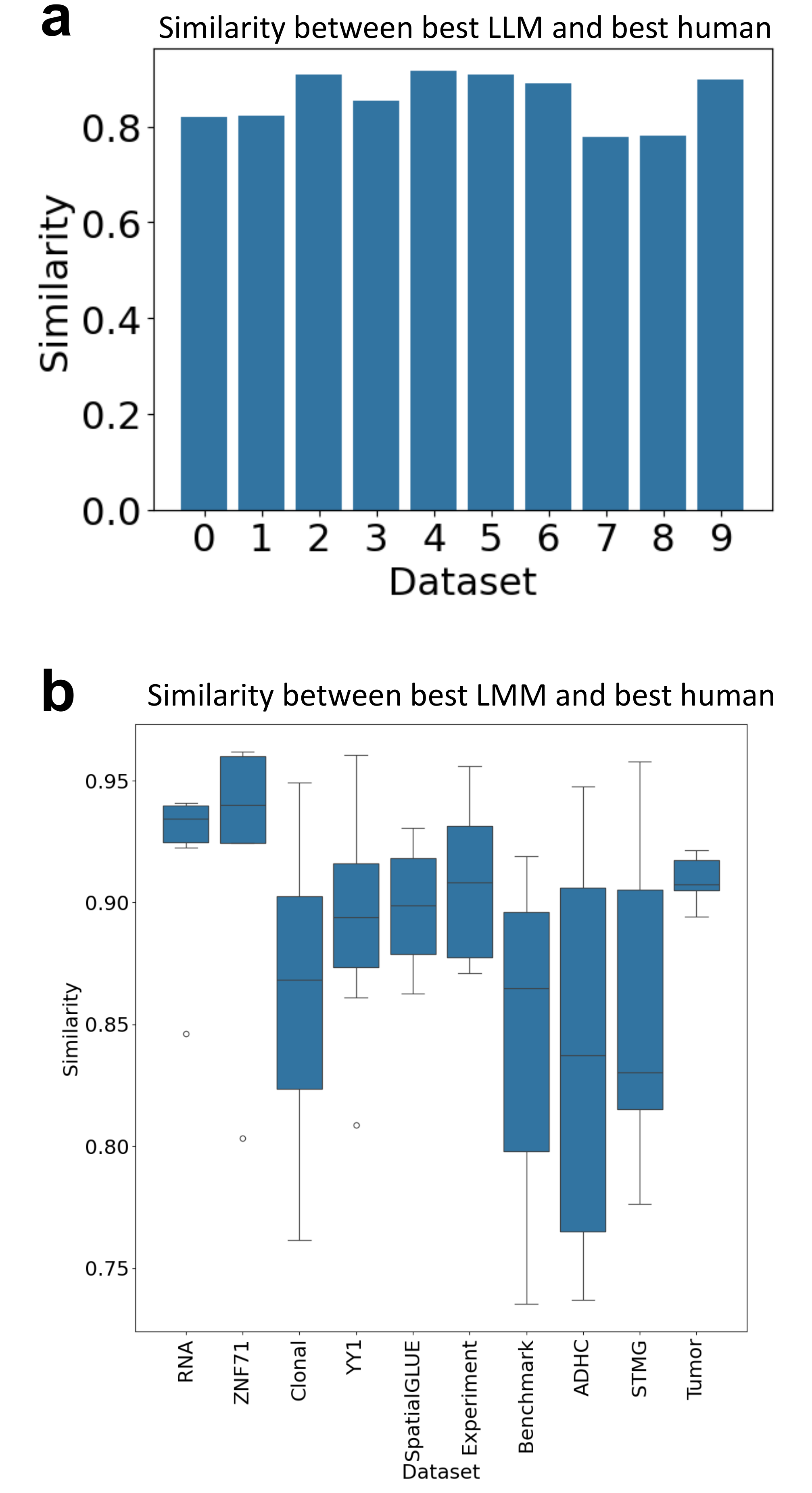}
    \caption{Similarity between human outputs and AI model outputs. (a) Similarity between outputs from the best LLM and the best human for text summarization. (b) Similarity between outputs from the best LLM and the best human for scientific figure understanding.}
    \label{supfig:sim info}
\end{figure}

\clearpage

\begin{figure}
    \centering
    \includegraphics[width=1\linewidth]{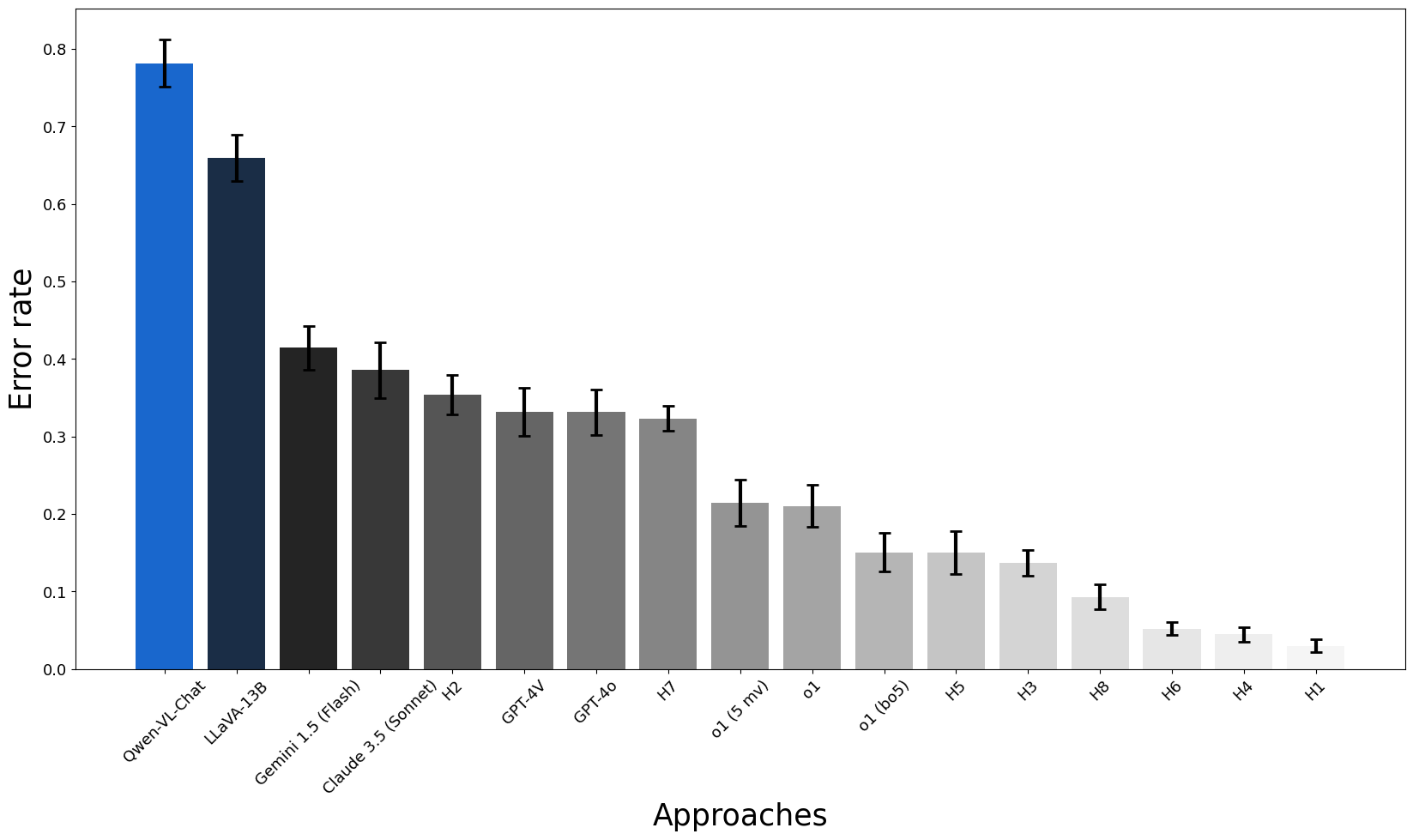}
    \caption{Analysis of error rate (cost of verification) for both LMMs and human experts.}
    \label{supfig:error rate analysis}
\end{figure}

\clearpage

\begin{figure}
    \centering
    \includegraphics[width=0.6\linewidth]{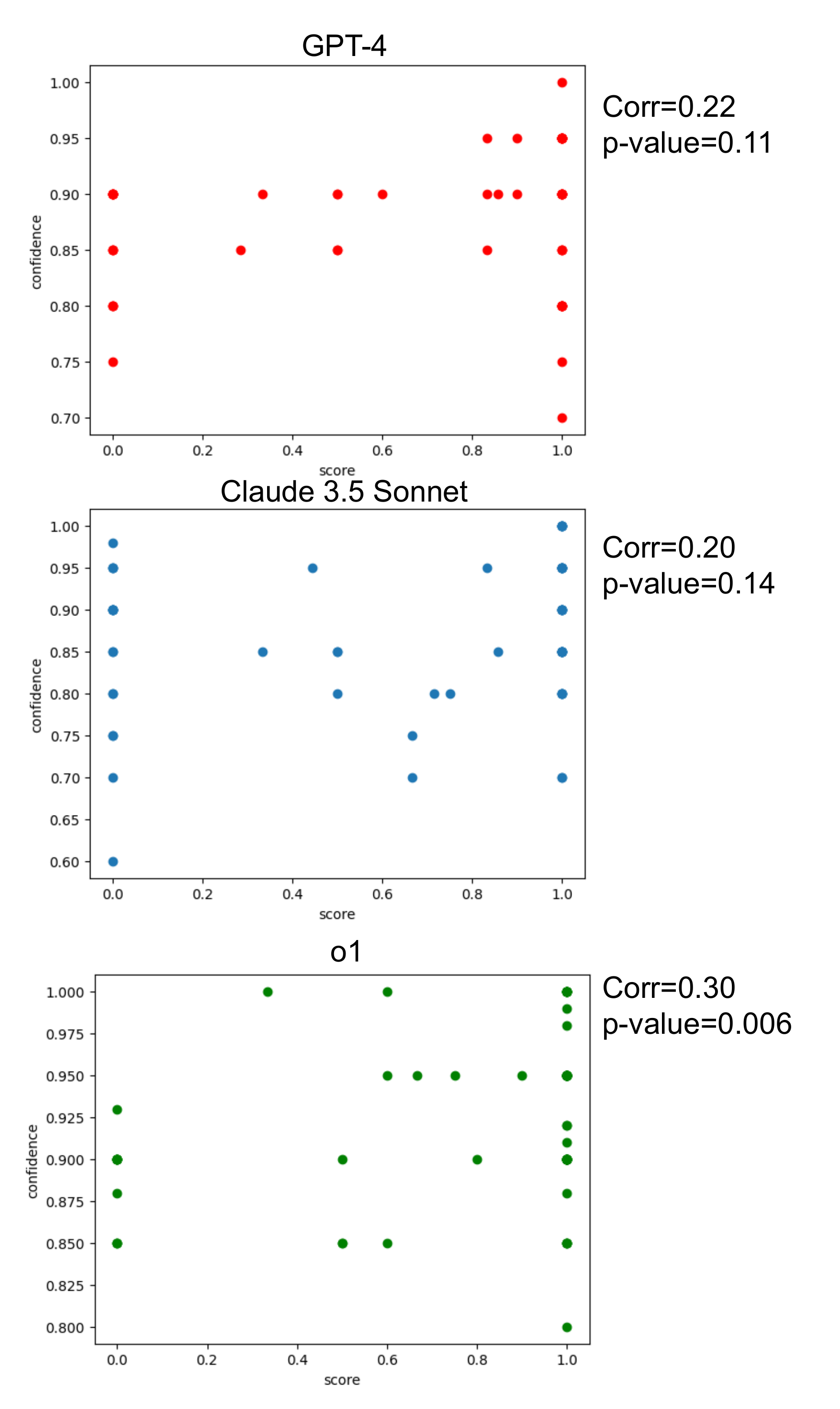}
    \caption{Confidence levels generated by different LMMs for scientific figure understanding.}
    \label{supfig: confidence plot}
\end{figure}

\begin{figure}
    \centering
    \includegraphics[width=1\linewidth]{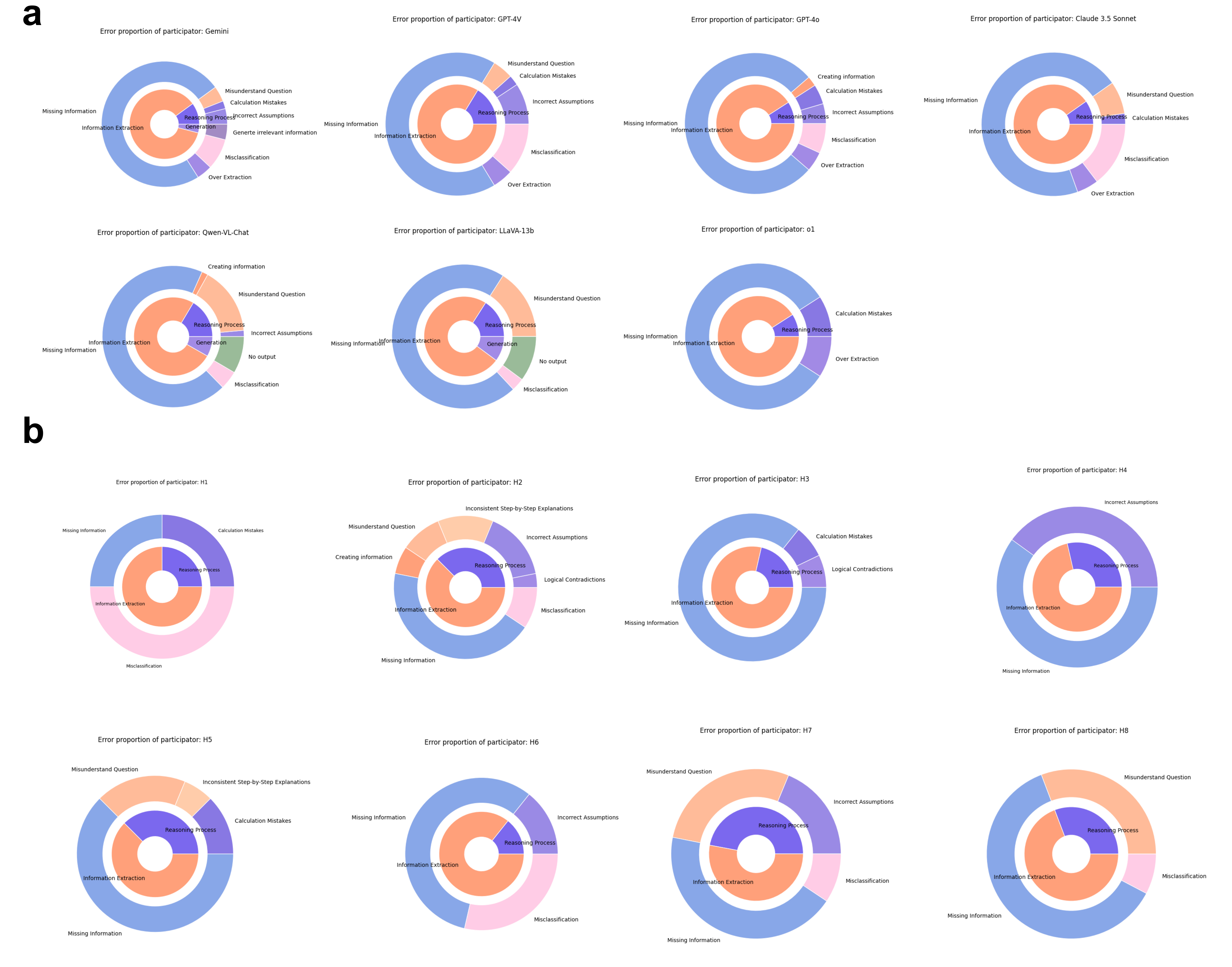}
    \caption{Information of error type analysis. (a) Proportion of errors across different LMMs. (b) Proportion of errors across different human participators.}
    \label{supfig:error type analysis}
\end{figure}

\begin{figure}
    \centering
    \includegraphics[width=1\linewidth]{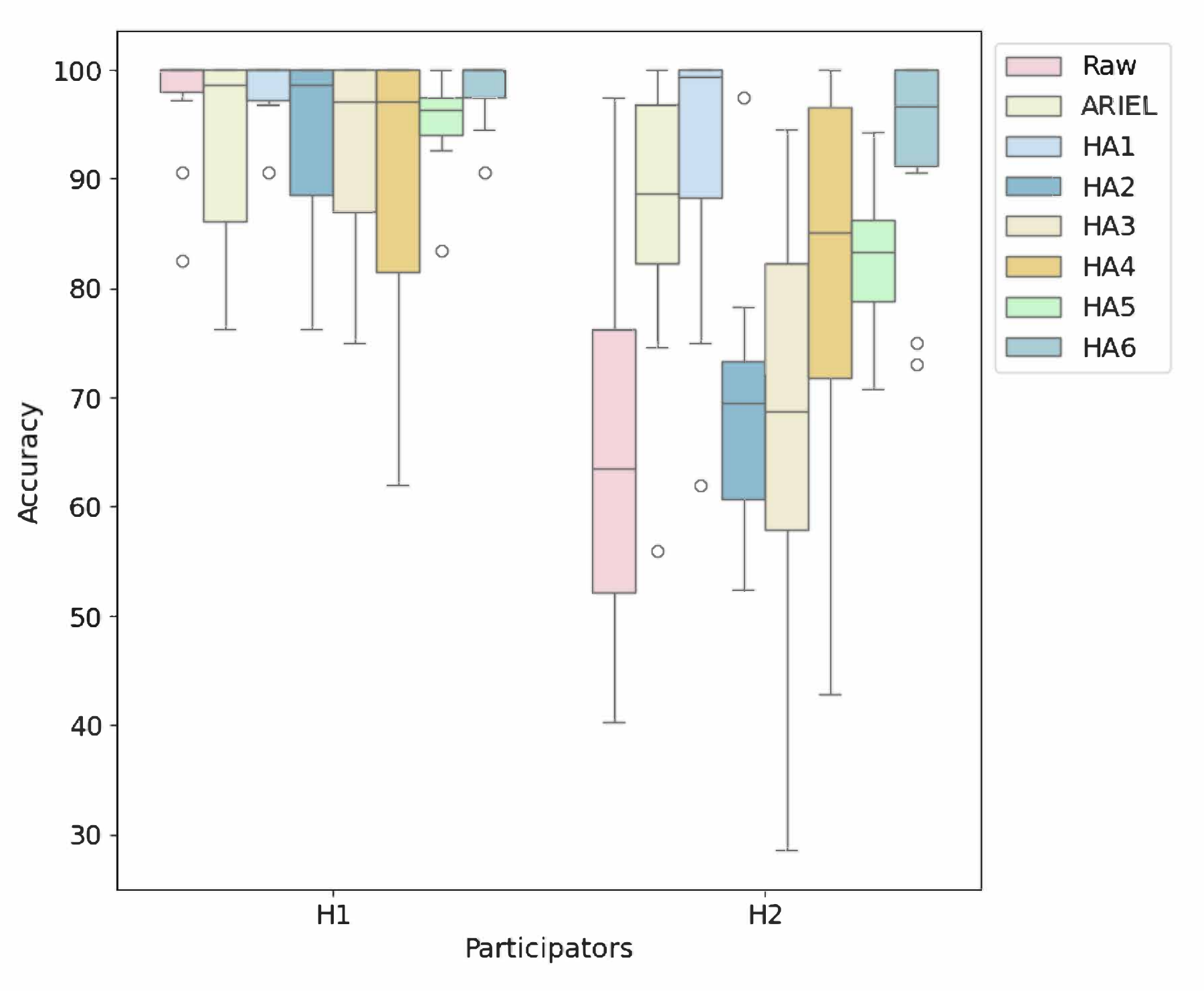}
    \caption{Boxplots with detailed metrics of comparing human assistants and LMM assistants for correcting answers (n$=$79 per group; triangle, mean; center line, median; box limits, upper and lower quartiles; whiskers, up to 1.5$\times$interquartile range; points, outliers).}
    \label{supfig:human correct details}
\end{figure}

\clearpage
\begin{figure}
    \centering
    \includegraphics[width=1\linewidth]{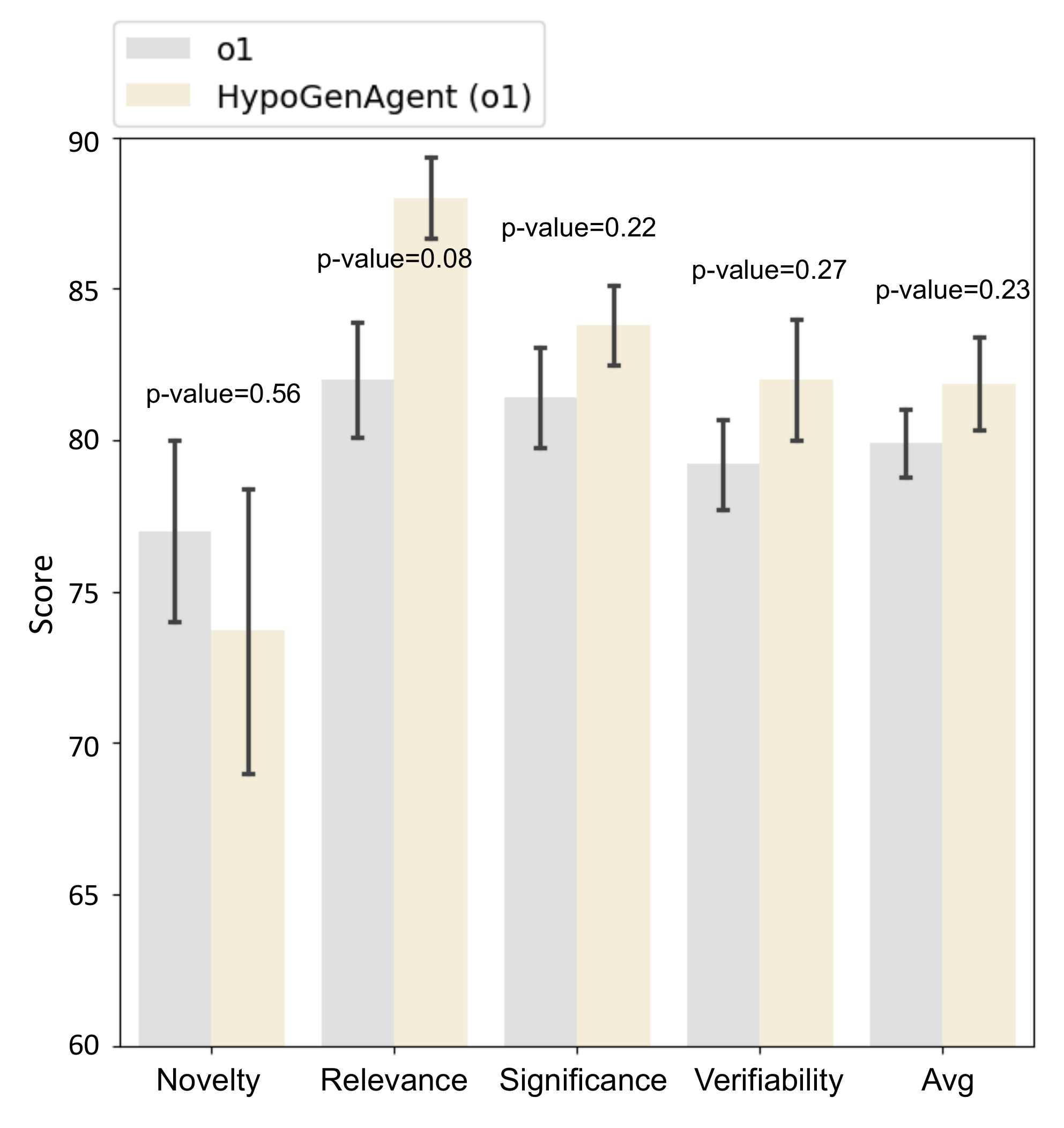}
    \caption{Hypothesis evaluation from four dimensions made by LLM as a judge. The p-value is computed with two-sided Mann-Whitney U test.}
    \label{supfig: llmasajudge_image}
\end{figure}

\clearpage
\begin{figure}
    \centering
    \includegraphics[width=1\linewidth]{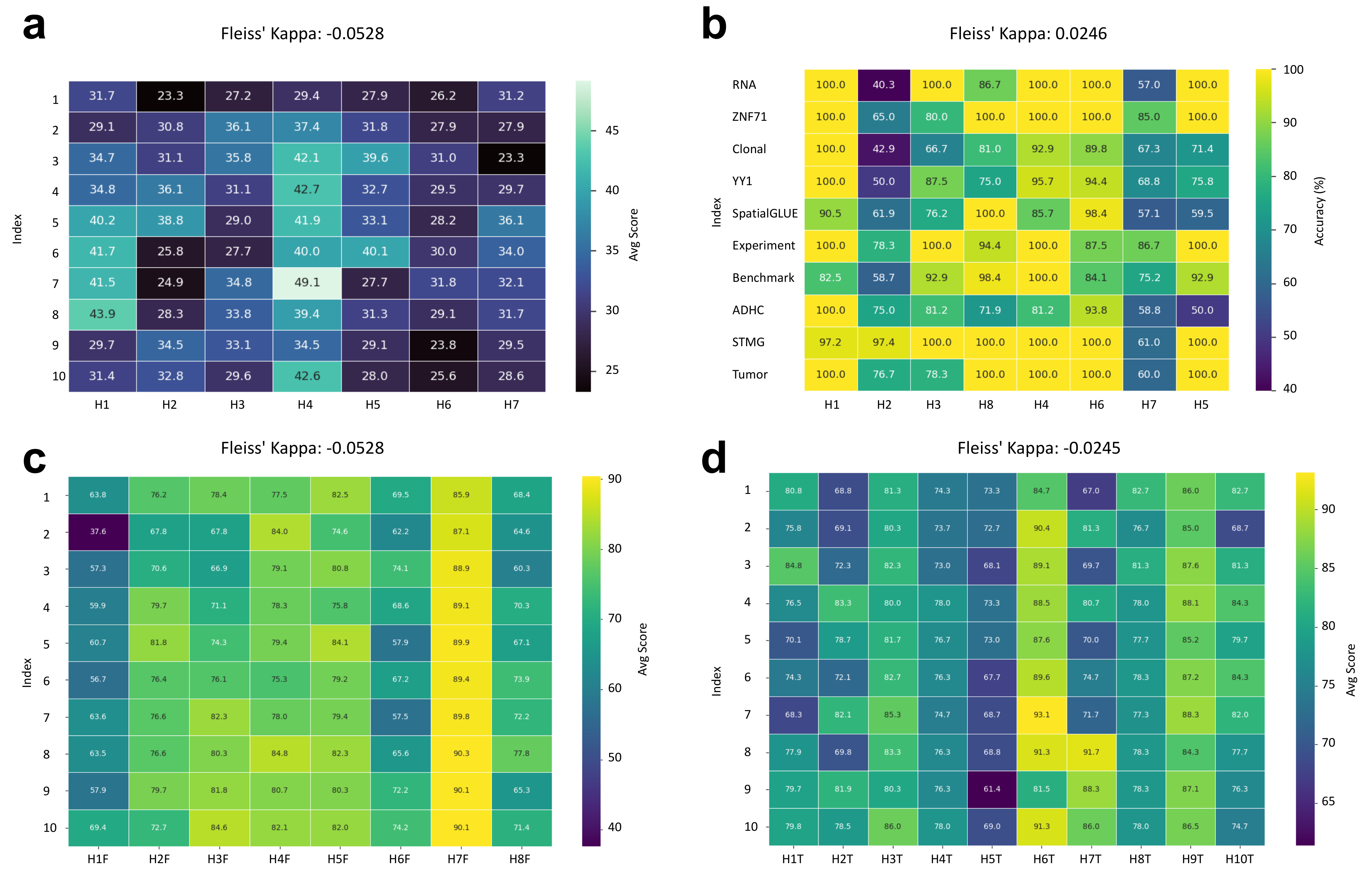}
    \caption{Analyzing the consistency across human experts. We report the Fleiss' Kappa score for each task. (a) Scores provided by different human experts in the text summarization task. (b) Scores provided by different human experts in the image understanding task. (c) Scores provided by different human experts in the novelty evaluation task. (d) Scores provided by different human experts in the summary quality evaluation task.}
    \label{supfig: consistency check}
\end{figure}

\begin{figure}
    \centering
    \includegraphics[width=1\linewidth]{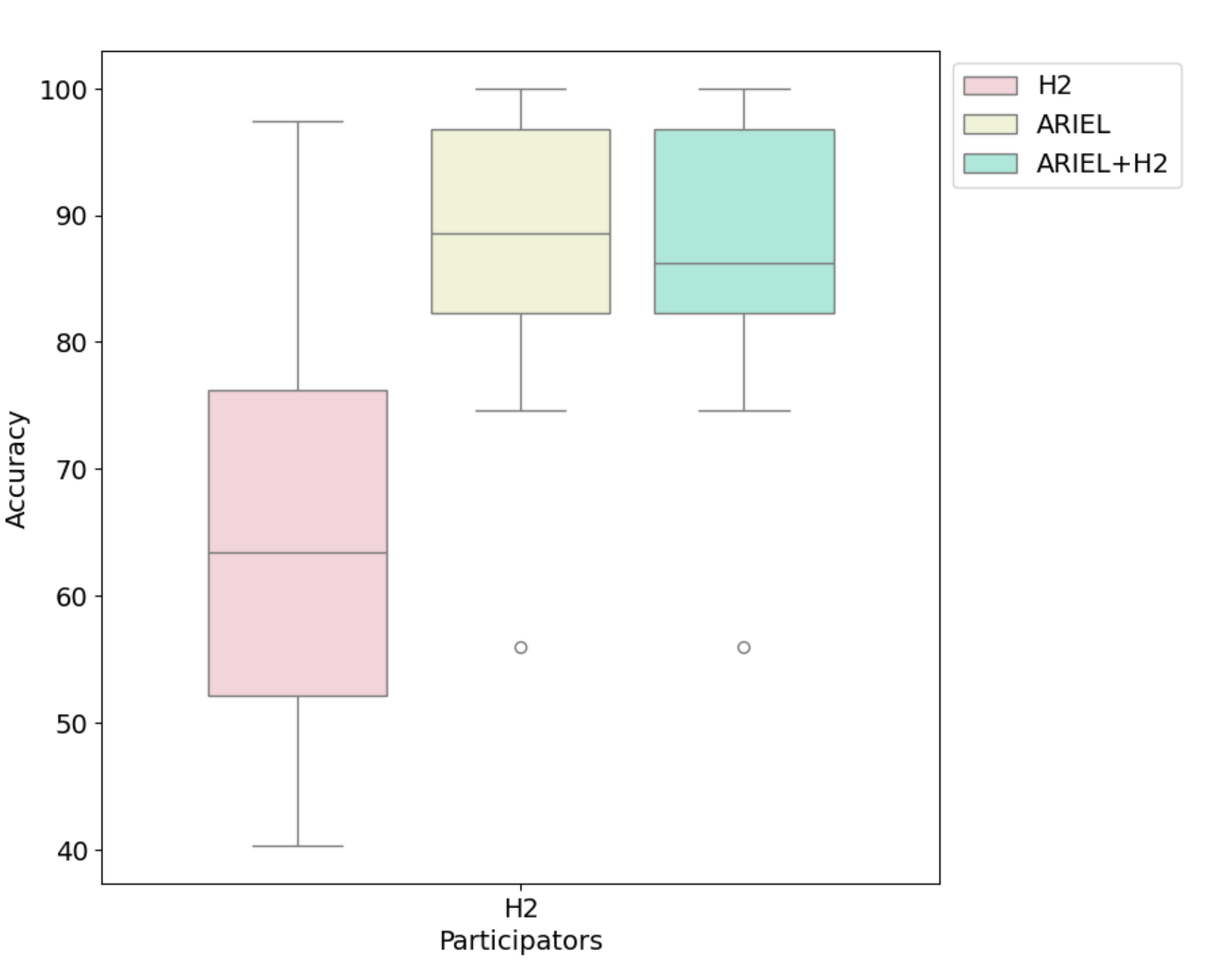}
    \caption{Boxplots with detailed metrics of comparing human assistants, LMM assistants, and human assistants using LMM for correcting answers (n$=$79 per group; triangle, mean; center line, median; box limits, upper and lower quartiles; whiskers, up to 1.5$\times$interquartile range; points, outliers).}
    \label{supfig:lmm include human correct details}
\end{figure}